\documentclass[11pt]{article}

\usepackage[preprint]{acl}

\usepackage{times}
\usepackage{latexsym}

\usepackage[T1]{fontenc}

\usepackage[utf8]{inputenc}

\usepackage{microtype}

\usepackage{inconsolata}

\usepackage{graphicx}
\usepackage{comment}
\usepackage{tcolorbox}
\usepackage{amsmath}
\usepackage{array}
\usepackage{amssymb}
\usepackage{booktabs}
\usepackage{multirow}
\usepackage{pifont}  
\usepackage{caption}
\usepackage{siunitx}
\usepackage{xcolor}
\usepackage{enumitem}
\usepackage{amssymb}
\usepackage{subcaption}

\usepackage{bbm}
\newcommand{\cmark}{\textcolor{green}{\ding{51}}} 
\newcommand{\xmark}{\textcolor{red}{\ding{55}}}   
\newcommand{\improvement}[1]{\textcolor[rgb]{0.0, 0.6, 0.0}{#1}}
\newcommand{\worsened}[1]{\textcolor{red}{#1}}
\title{Style Amnesia: Investigating Speaking Style Degradation and Mitigation \\in Multi-Turn Spoken Language Models}

\author{Yu-Xiang Lin$^1$, Cheng-Han Chiang$^1$, Hung-yi Lee$^{1,2}$ \\
        $^1$National Taiwan University \\
        $^2$NTU Artificial Intelligence Center of Research Excellence (NTU AI-CoRE)}

\begin{document}
\maketitle
\begin{abstract}


In this paper, we show that when spoken language models (SLMs) are instructed to speak in a specific speaking style at the beginning of a multi-turn conversation, they cannot maintain the required speaking styles after several turns of interaction; we refer to this as the~\textit{style amnesia} of SLMs. 
We focus on paralinguistic speaking styles, including emotion, accent, volume, and speaking speed. 
We evaluate three proprietary and two open-source SLMs, demonstrating that none of these models can maintain a consistent speaking style when instructed to do so. 
We further show that while SLMs can recall the style instruction when prompted in later turns, they still fail to express it, but through explicit recall can mitigate~\textit{style amnesia}. 
In addition, SLMs struggle more when the style instruction is placed in system messages rather than user messages, even though system messages are specifically designed to provide persistent, conversation-level instructions. 
Our findings highlight a systematic gap in current SLMs' ability to maintain speaking styles, highlighting the need for improved style adherence in future models. Our code and evaluation data are publicly available at \url{https://github.com/YuXiangLin1234/SLM-Style-Amnesia}.

\end{abstract}

\section{Introduction}

Spoken language models (SLMs) can take speech input and generate speech responses.
Unlike text-only large language models (LLMs), SLMs integrate audio encoders and vocoders~\cite{kong2020hifi} to support end-to-end (E2E) speech understanding and generation~\citep{defossez2024moshi, zeng2024glm, fangllama, wu2025step, huang2025step}.
While this E2E design introduces additional challenges beyond text processing, such as handling paralinguistic features and non-textual information, current SLMs have demonstrated the capability to detect user attributes like emotion, accent, and gender.
By leveraging these acoustic cues, SLMs can adjust their outputs to produce more contextually appropriate responses~\citep{wu2025step, 
GeminiLiveUpdate2025}.
Beyond passive perception, SLMs are also capable of actively following the user-specified speaking style in single-turn interactions~\citep{zeng2024glm, wu2025step, chiang2025audio, zhan2025vstyle}.

\begin{figure}[t!]
    \centering
    \includegraphics[width=0.9\columnwidth]{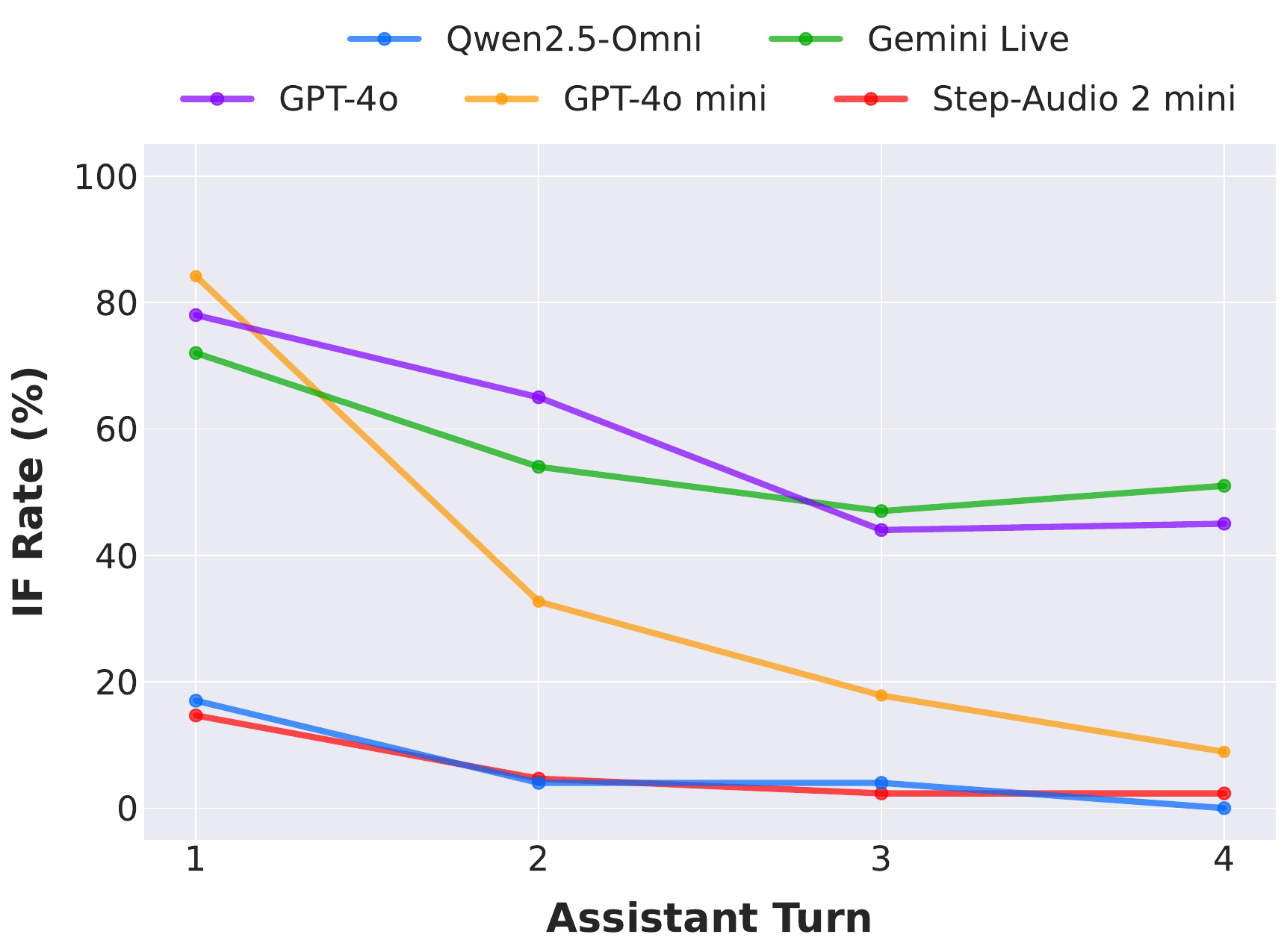}  
    \caption{When instructed to consistently speak sadly throughout the conversation, SLMs try their best to follow the instruction in the first turn, but the instruction-following rate degrades rapidly in subsequent turns.}
    \label{fig:sadness_ifrate}
\end{figure}

Despite these advances, most prior works focus on \textit{single-turn} evaluation when evaluating the expressive speech generation of SLMs.
It is thus unclear whether SLMs can consistently follow a user-specified speaking style in a \textit{multi-turn} spoken interaction.
To address this gap, we investigate the speaking style consistency of SLMs in multi-turn spoken conversations.
Precisely, we instruct an SLM we want to evaluate to ``\textit{follow a specific \textbf{speaking style} throughout the conversation}'' at the beginning of the dialogue and then interact with the SLM in a multi-turn conversation.
The speaking style can be emotion, accent, volume, or speaking speed.
After collecting model responses at each turn, we assess the style instruction-following (IF) rate using style-specific automatic judges.

Our results show that the speaking style instruction-following rate degrades over interaction turns, a phenomenon we refer to as \textit{style amnesia}.
Figure~\ref{fig:sadness_ifrate} presents an example when the SLM is instructed to consistently speak sadly, but no SLM can consistently do this.
Importantly, we find that most models exhibit a higher style IF rate in the first turn, while the IF rate gradually declines over subsequent turns.
These findings indicate that maintaining consistent speaking styles over multiple turns remains challenging for current SLMs.
We conduct comprehensive experiments to understand~\textit{style amnesia} and explore potential methods to resolve this issue.

Our contribution can be summarized as follows:
\begin{itemize}[leftmargin=16pt, itemsep=0pt, topsep=3pt, after=\vspace{3pt}, labelsep=0.5em]
    \item We identify that in multi-turn conversations, SLMs fail to consistently follow the given style instruction given in the first turn.
    \item We show that explicitly prompting SLMs to recall the initial style instruction can alleviate~\textit{style amnesia}.
    \item We comprehensively analyze the mechanisms behind~\textit{style amnesia} by tracking attention dynamics, evaluating the impact of prompt placement, and assessing text-acoustic synergy.
\end{itemize}

\section{Related Works}

\subsection{Context Loss in Multi-Turn Dialogue}

Although LLMs support multi-turn interactions, recent studies have shown that their performance often degrades as dialogues become longer~\cite{kwan-etal-2024-mt, han-etal-2025-language, qamar-etal-2025-llms, li-etal-2025-structflowbench,  laban2025llms}.
These studies indicate the shortcomings of current LLMs, such as difficulties with multi-turn instruction following~\cite{han-etal-2025-language, li-etal-2025-structflowbench}, loss of information in long dialogue~\cite{kwan-etal-2024-mt, qamar-etal-2025-llms}, and errors caused by fragmented information across turns~\cite{laban2025llms}.

In terms of SLMs' evaluation, SpokenWOZ~\cite{si2023spokenwoz} highlights the difficulty of aggregating fragmented information across turns.
$C^3$~\cite{ma2025c3} focuses on bilingual spoken dialogue and evaluates models' capabilities in handling complex multi-turn conversational phenomena, including omission and coreference.
ContextDialog~\cite{kim-etal-2025-voice-assistant}, on the other hand, demonstrates that SLMs exhibit significantly lower memorization performance in long dialogue compared to their text-based counterparts.
Overall, these works assess the ability of SLMs to retain and utilize information in long spoken dialogues but leave the expressiveness of generated speech unevaluated.

\begin{table}[t!]
    \centering
    \resizebox{\columnwidth}{!}{
        \begin{tabular}{lcccc}
        \toprule
         & Multi-Turn & Style Eval. & Interactive & Turn Analysis \\
        \midrule
        SpokenWOZ~\cite{si2023spokenwoz}     & \cmark & \xmark & \xmark & \xmark \\
        C$^3$~\cite{ma2025c3}         & \cmark & \xmark & \xmark & \xmark \\  
        ContextDialog~\cite{kim-etal-2025-voice-assistant} & \cmark & \xmark & \xmark & \xmark \\
        Vstyle~\cite{zhan2025vstyle}        & \xmark & \cmark & \xmark & \xmark \\
        Game-Time~\cite{chang2025game}     & \xmark & \cmark & \xmark & \xmark \\
        StyleSet~\cite{chiang2025audio} & \xmark & \cmark & \xmark & \xmark \\
        URO-Bench~\cite{yan-etal-2025-uro}     & \cmark & \cmark & \xmark & \xmark \\
        VocalBench~\cite{liu2025vocalbench}    & \cmark & \cmark & \xmark & \xmark \\
        VoxDialogue~\cite{cheng2025voxdialogue}   & \cmark & \cmark & \xmark & \xmark \\
        Multi-Bench~\cite{deng2025multi}   & \cmark & \cmark & \cmark & \xmark \\
        \midrule
        \textbf{Ours} & \cmark & \cmark & \cmark & \cmark \\
        \bottomrule
        \end{tabular}
    }
    \caption{Comparison with related works.~\textit{Style Eval.} refers to the assessment of speaking style.}
    \label{tab:prior_work_comparision}
\end{table}

\subsection{Speaking Style Assessment}

Beyond basic intelligibility and audio quality evaluation~\cite{saeki2022utmos, fangllama}, a growing body of work has started to focus on the expressiveness of SLMs~\cite{zeng2024glm, wu2025step}.
For example, Vstyle~\cite{zhan2025vstyle} evaluates whether SLMs can adapt their voice style according to the given instruction. Game-Time~\cite{chang2025game} focuses on the temporal control of SLMs.
Besides,~\citet{chiang2025audio} demonstrate that Large Audio-Language Models (LALMs) are capable of serving as an automatic judge to evaluate the voice style. 
Their experiment shows high agreement between human evaluation results and those from Gemini-2.5 Pro~\cite{comanici2025gemini}. 

Although previous works have evaluated the expressiveness of SLMs, voice style consistency in multi-turn interactions remains largely underexplored.
URO-bench~\cite{yan-etal-2025-uro}, VocalBench~\cite{liu2025vocalbench}, and VoxDialogue~\cite{cheng2025voxdialogue} use predefined dialogues as the first $k-1$ turns and ask SLMs to generate the $k$-th response, which does not support turn-level analysis. 
In contrast to approaches based on predefined multi-turn dialogues, we adopt a model-based user simulator to interact with the evaluated SLMs.
We argue that this setting more closely reflects real-world conversational scenarios.

A concurrent work, Multi-Bench~\cite{deng2025multi}, also evaluates SLMs in multi-turn interactive dialogues.
However, their evaluation aggregates performance into a single global score to assess overall emotional intelligence.
In contrast, our turn-level analysis provides fine-grained and informative insights into how inconsistencies emerge across a wide range of speaking styles, including emotion, accent, speed, and volume.
A comparison with prior work is summarized in Table~\ref{tab:prior_work_comparision}.


\section{Evaluating Multi-Turn Speaking Style Following of SLMs}

We aim to evaluate whether SLMs can follow a speaking style instruction given at the beginning of a dialogue throughout the whole conversation.
In this section, we first introduce our motivation in Section~\ref{subsection: motivation}, and then describe the evaluation framework, dataset, and evaluation metrics in the following subsections.

\subsection{Motivation}
\label{subsection: motivation}

In the real world, each user may have different preferences for their SLM's speaking style.
Consequently, it is important that an SLM can follow the speaking style instruction specified by the user.
The user may specify some style instructions for the SLM and expect it to follow them throughout the dialogue.
While SLMs are capable of following style instructions in \textit{single-turn} interaction, it is unrealistic to expect the user to restate the same style instruction in each round of the interaction.
Thus, whether an SLM can consistently follow a speaking style given at the beginning of the conversation is an important ability of SLMs.
Next, we introduce our evaluation framework to evaluate this ability.

\subsection{Evaluation Framework}
Given an SLM, our goal is to provide it with \textbf{speaking style instructions} and ask it to adhere to this style throughout the \textbf{multi-turn conversation} based on a specified topic.
We then evaluate each turn of the SLM-generated response.
An illustration of the overall framework is shown in Figure~\ref{fig:framework}.
In this section, we introduce three important components in our framework: (1) the speaking style instructions given to the SLM, (2) the conversation topic, and (3) how to interact with the SLM to form a multi-turn dialogue with a user simulator.

\subsubsection{Speaking Style Instructions}
\label{subsubsection: Speaking Style Instructions}
At the beginning of the dialogue, we will instruct the SLM to follow a specific speaking style. 
Unless otherwise specified, we give the instruction in the first user turn, as shown in Figure~\ref{fig:framework}.
In our paper, we focus on paralinguistic speaking styles, as this is what makes SLMs different from text-only LLMs.
We include four types of paralinguistic attributes, each with multiple possible values.
The included speaking styles are listed as follows:
\begin{itemize}[leftmargin=12pt, itemsep=0pt, topsep=5pt, after=\vspace{4pt}, labelsep=0.5em]
    \item \textbf{Emotion}: Sad, happy, angry, or neutral tone. 
    \item \textbf{Accent}: North American or Indian accents. 
    \item \textbf{Volume}: A higher or lower volume. 
    \item \textbf{Speed}: A faster or slower pace.
\end{itemize}

Among many possible paralinguistic speaking styles, we select the above attributes and values since they can be automatically evaluated, as detailed in Section~\ref{sec:evaluation_metrics}.

\subsubsection{Dialogue Topics}
\label{subsubsection: Dialogue Topics}
\begin{figure}
    \centering
    \includegraphics[width=\linewidth]{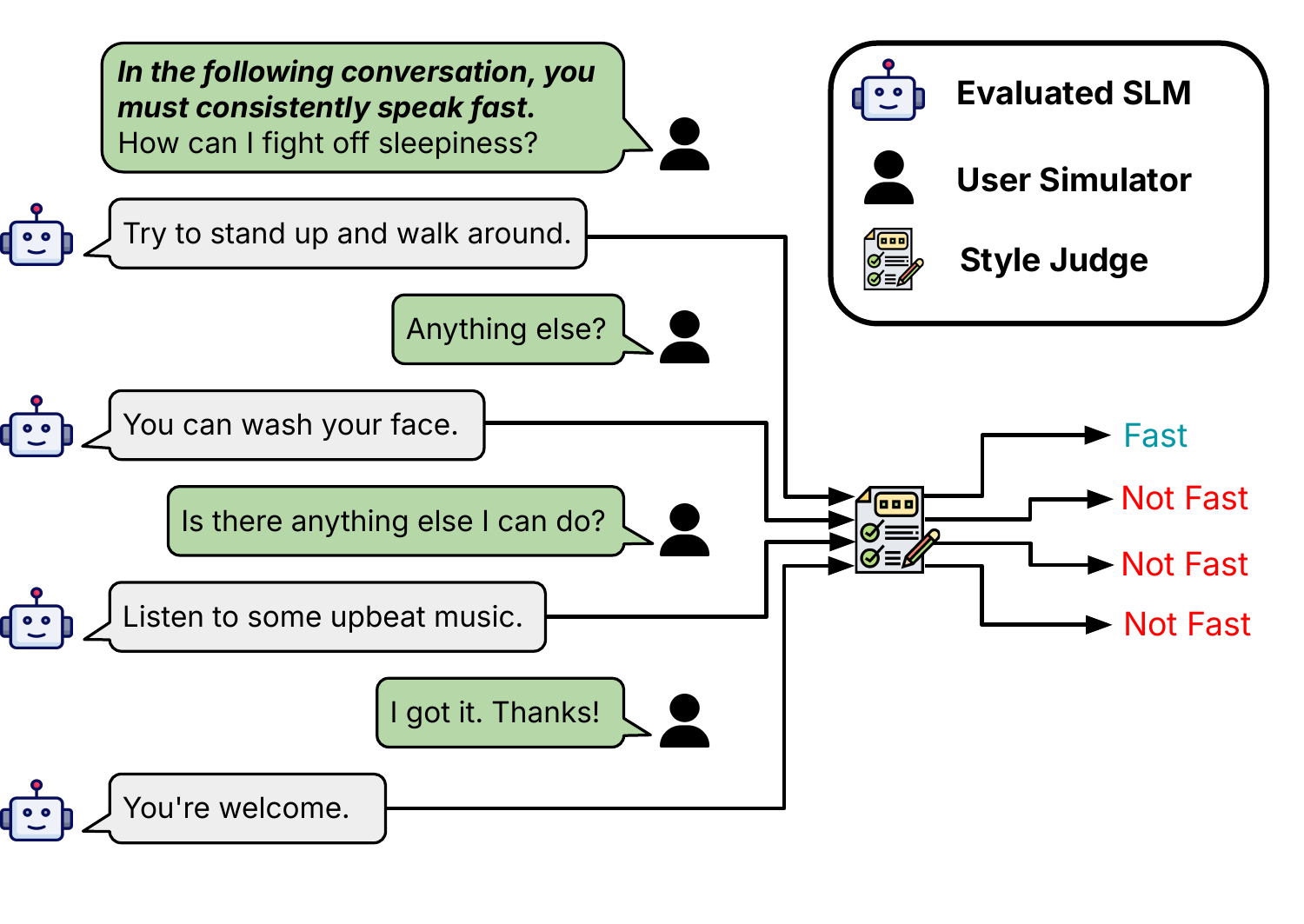}
    \vspace{-20px}
    \caption{The overview of the evaluation framework.
    The SLM can speak fast in the first turn as instructed, but it fails to maintain this style in later dialogue turns.}
    \label{fig:framework}
\end{figure}

In the first user turn of the dialogue, we provide a conversation opener for each dialogue.
The conversation opener is used to control the topic that the SLM should talk about during each evaluation, ensuring that different SLMs are evaluated under a common conversational setting.
We select the topic from Soda~\cite{kim2023soda}, a large-scale dataset of social interaction dialogues.
The details about how we select the topics from Soda are shown in Appendix~\ref{app:dataset-construction}.
Eventually, we collect 100 diverse conversation openers to initiate the conversations.
We manually inspect the selected topics and filter out improper topics, such as queries about personal preferences that are unsuitable for machine conversations.
An example conversation opener is shown in Figure~\ref{fig:framework} ``How can I fight off sleepiness?''

\subsubsection{Multi-Turn Interaction Using a User Simulator}
\label{subsubsection: Multi-Turn Interaction}
At the core of our evaluation framework is the multi-turn interaction between the evaluated SLM and a user.
To enable back-and-forth interactions with an evaluated SLM, we build a \textit{user simulator} with a cascade SLM.
The cascade SLM is a speech-in-speech-out system composed of an ASR, a text-only LLM, and a TTS model.
The ASR model takes the speech input from the evaluated SLM and transcribes it into text.\footnote{Most existing SLMs output both speech and its corresponding transcription at the same time. In this case, we directly use the text output from the evaluated SLM and skip the ASR module in the user simulator.}
A text-only LLM, powered by GPT-5 mini~\cite{OpenAIGPT5mini2024}, takes the text input and generates a text response based on the transcription produced by the evaluated SLM. 
Last, the TTS model, GPT-4o mini TTS~\cite{OpenAIGPT4oMini2024}, converts the text output from the LLM into speech and sends it back to the evaluated SLM to continue the conversation.

\subsubsection{Summary}

By combining 10 speaking styles and 100 topics, each SLM is evaluated with \textbf{1,000} dialogues.
Currently, we do not compose multiple types of speaking styles in a single instruction; this is mainly because SLMs cannot even properly follow a single speaking style instruction consistently, as later shown in Section~\ref{section: Main exp}. 
However, our evaluation framework can be easily extended to the composition of multiple types of speaking styles, and we leave this as future work.

\subsection{Evaluation Metrics}
\label{sec:evaluation_metrics}

To quantify how well SLMs adhere to the target speaking style, we adopt the ~\textbf{instruction-following (IF) rate}, denoted as $IF$, as our primary evaluation metric. 
Given a target speaking style $s$ from Section~\ref{subsubsection: Speaking Style Instructions}, e.g., \textit{speak sadly}, or \textit{speak fast}, etc., and dialogue topic $i$, the evaluated SLM generates output $o_{i,j}$ at turn $j$, the IF rate for style $s$ at turn $j$ denoted as $IF_{j}(s)$ and defined by:
\begin{equation}
\label{eq:ifrate}
IF_j(s) = \frac{\sum_{i=1}^{N}\mathbbm{1}(o_{i,j}, s)}{N} \times 100\%.
\end{equation}
Here, $N=100$ is the total number of topics, and $\mathbbm{1}(\cdot)$ is a binary indicator determined by a style-specific judge (described later in Section~\ref{subsubsection: Automatic Judges for Styles}), which reflects whether the generated output $o_{i,j}$ follows the given style instruction $s$.
This metric indicates the percentage of generated speech that correctly follows the given style instruction.

Our goal is to quantify how well the style IF ability changes over the conversation turns.
To do so, we define two metrics to measure how good the style IF is across multiple interaction turns.

\textbf{(1) First-turn IF rate} $IF_1$. \quad
This metric serves as a reference to evaluate the capability of the SLM to follow the specific style in the first turn.
If an SLM fails to achieve a reasonable $IF_1$, we may not expect it to perform better in later turns.

\textbf{(2) Degradation rate} $D$. \quad
This metric quantifies how the IF rate decays over subsequent turns relative to the initial performance $IF_1$.
It is defined as the average absolute difference between the instruction-following rates of subsequent SLM turns and that of the first SLM turn.

\begin{equation}
    \label{eq:degradation_rate}
    D = \sum_{j=2}^K \frac{\max(IF_1(s) - IF_j(s), 0)}{K-1}.
\end{equation}

\noindent Here $K$ denotes the total number of assistant turns of the evaluated SLM.
We apply the $\max(\cdot, 0)$ operator to focus solely on style degradation, thereby avoiding the cancellation of positive and negative differences during averaging.
For $IF_1$, larger values indicate better first-turn control over the requested voice style.
A smaller $D$ indicates less style degradation.
In this paper, we set $K = 4$.

Aside from measuring the speaking style, we also measure the semantic coherence of the dialogue to ensure that the content is valid using LLM-as-a-judge~\citep{chiang-lee-2023-large}.
Since the semantic coherence of the dialogue is not our primary focus, we leave the details in Appendix~\ref{app:dialogue_conherence_evaluation}.
Overall, all SLMs we evaluate maintain reasonable semantic consistency across turns.

\begin{figure*}[t!]
    \centering
    \includegraphics[width=0.99\linewidth]{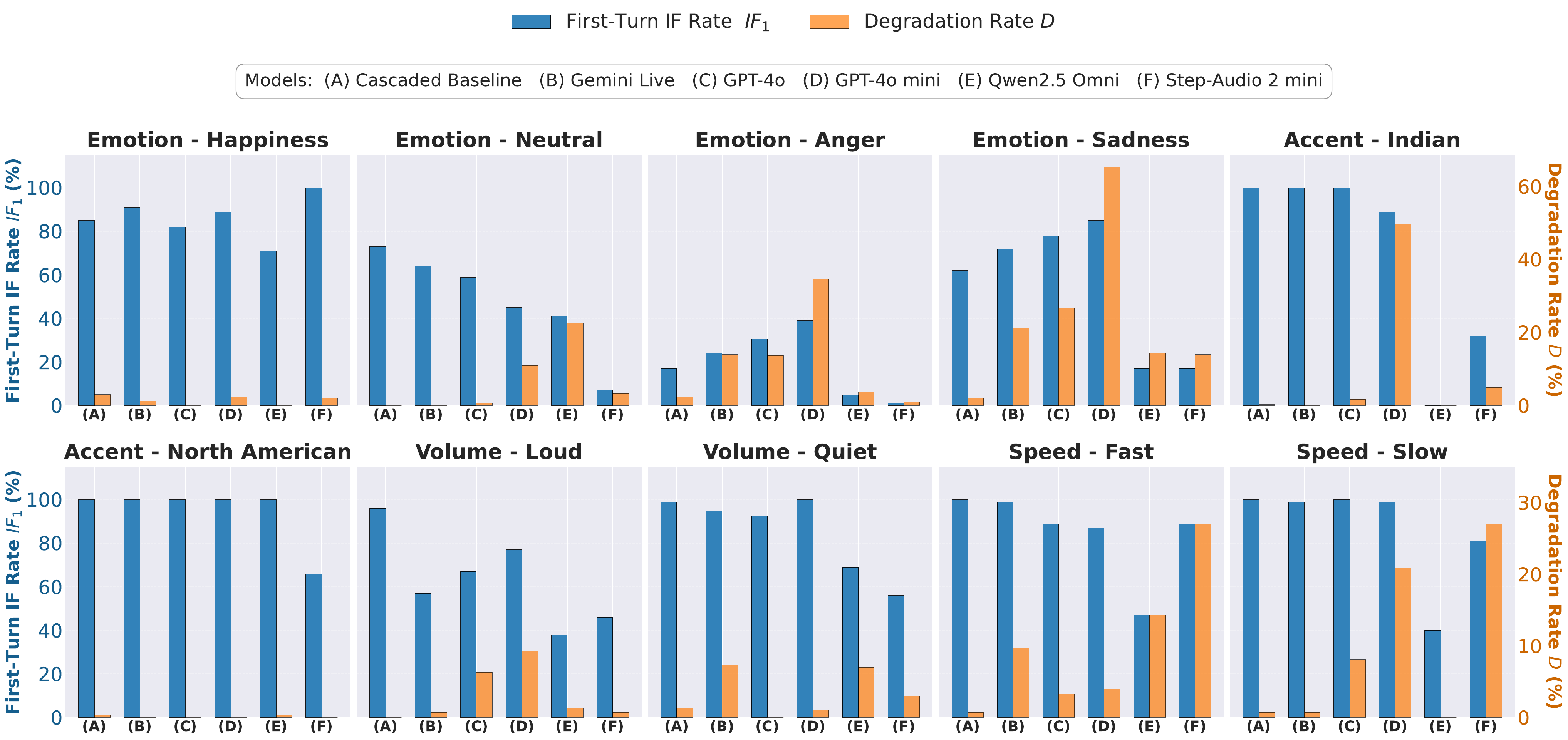}
    \caption{The first-turn IF rate $IF_1$ and degradation rate $D$ across different speaking styles.}
    \label{fig:main_result}
\end{figure*}

\subsubsection{Automatic Judges for Styles}
\label{subsubsection: Automatic Judges for Styles}

An automatic style judge takes the target speaking style $s$ and a speech input $o_{i,j}$ and returns 0 or 1 to indicate whether the $o_{i,j}$ aligns with the target style $s$.
Different automatic judges are adopted for different types of speaking styles.

\textbf{Emotion}\quad
We use Emotion2vec-Large~\cite{ma2024emotion2vec} to predict the emotion of $o_{i,j}$. 
Although it is a 9-class model, we focus on the probabilities for happiness, sadness, anger, and neutral. 
We take the maximum probability among these four categories as the final prediction and verify whether it aligns with $s$.
If the prediction is aligned, the indicator function $\mathbbm{1}(\cdot)$ is 1; otherwise, it is 0. 

\textbf{Accent}\quad We use Voxlect-English-Dialect-Whisper-Large-v3~\cite{feng2025voxlect}  to predict the accent of $o_{i,j}$.
From this 16-class model, we extract the probabilities for North American and Indian English, taking the maximum as the predicted accent of $o_{i,j}$.
If it is aligned with $s$, $\mathbbm{1}(\cdot)$ is 1; otherwise $\mathbbm{1}(\cdot)$ is 0.

\textbf{Volume}\quad We measure volume using Loudness Units Full Scale (\textit{LUFS}) with PyLoudnorm~\cite{steinmetz2021pyloudnorm}. 
Since subjective terms like ``\textit{loud}'' or ``\textit{quiet}'' lack universal definitions, we adopt a relative evaluation standard. 
We first establish a baseline by querying the SLM with a neutral instruction: ``\textit{You are a text-to-speech model. Please read the given text at a normal volume without adding or omitting anything},'' and generate another speech $o'_{i,j}$.
$\mathbbm{1}(\cdot)$ is 1 if the \textit{LUFS} of $o_{i,j}$ is higher (for~\textit{loud} instruction) or lower (for~\textit{quiet} instruction) than $o'_{i,j}$; otherwise $\mathbbm{1}(\cdot)$ is 0.

\textbf{Speed}\quad We measure speaking rate using Words Per Minute (\textit{WPM}) with Parakeet TDT v2~\cite{nvidiaasr2025}. 
Similar to volume, we compare the SLM generated speech $o_{i,j}$ with the TTS-generated speech $o'_{i,j}$ using the same model to determine whether $o_{i,j}$ is faster (or slower) than $o'_{i,j}$.

\section{Main Experiments}
\label{section: Main exp}
\subsection{Experimental Setup}

We evaluate three proprietary SLMs: GPT-4o\footnote{gpt-4o-audio-preview-2025-06-03}\,\cite{openai2024gpt4ocard}, GPT-4o mini\footnote{gpt-4o mini-audio-preview-2024-12-17}\,\cite{OpenAIGPT4oMini2024}, Gemini Live\footnote{gemini-2.5-flash-native-audio-preview-09-2025}\,\cite{GeminiLiveUpdate2025}, and two open-source end-to-end SLMs: Qwen2.5-Omni~\cite{xu2025qwen2} and Step-Audio 2 mini~\cite{wu2025step}.
We select these two representative open-source SLMs because they provide official vLLM implementations~\cite{kwon2023efficient} to speed up the inference process.

Apart from these E2E SLMs, we introduce a cascaded SLM baseline comprised of GPT-5 mini and Gemini-TTS~\cite{geminitts2025}.
The TTS in the baseline receives the style instruction in each turn.
The performance of the cascaded baseline serves as the performance upper-bound.
The hyperparameter choices of each model are provided in Appendix~\ref{app:hyperparameters}.

\subsection{Results}
\label{sec:main_results}
We report the first-turn IF rate $IF_1$ and the degradation rate $D$ in Figure~\ref{fig:main_result}.\footnote{GPT-4o and GPT-4o mini sometimes return only text transcription without speech. When this occurs, we query the models up to three times, but a few samples still fail.
We discuss this issue in Appendix~\ref{app:gpt_failure}. In the following section, all metrics are computed only on successful cases.} 
We also provide the illustration of~\textit{style amnesia} in Figures~\ref{fig:sadness_ifrate} and~\ref{fig:turn_level_visulization}, and IF rate for each turn in Appendix~\ref{app:full_degradation}.

For the~\textbf{Emotion} speaking style, the Cascaded Baseline demonstrates consistent IF rate across turns, with degradation within $3.0\%$.
In contrast, Gemini Live and GPT-4o show a degradation rate ranging from 13.7\% to 26.7\% for Anger and Sadness.
GPT-4o mini exhibits even larger degradation, reaching 34.7\% and 65.3\% for Anger and Sadness, respectively.
Qwen2.5-Omni and Step-Audio 2 mini also show around 14.0\% degradation for Sadness.
However, their degradation rates for Anger are only 3.7\% and 1.0\%, because both models already struggle to perform this style effectively in the first turn.

For the~\textbf{Accent} speaking style, Gemini Live shows stable performance when maintaining an Indian English accent, suggesting strong control over this attribute.
In contrast, GPT-4o mini still exhibits nearly 50\% degradation for the Indian English accent.
Step-Audio 2 mini achieves a 32.0\%  $IF_1$ but a 5.0\% degradation rate.
Qwen2.5-Omni performs well with the North American accent but fails to produce the Indian accent. 

Interestingly, we observe that nearly all SLMs perform better when generating happiness, neutral tone, and North American English than other style attributes.
We hypothesize that this discrepancy arises because these attributes correspond to the default speaking styles of the evaluated SLMs.
To validate this assumption, we analyze the emotion and accent distributions of samples generated during the speed and volume evaluations, as these prosodic features can coexist with emotion and accent.
The distributions shown in Appendix~\ref{app:default_style} support our hypothesis: most SLMs tend to default to happy or neutral tones and North American accents, thereby leading to a low degradation rate. 

\begin{figure}[t]
    \centering
    \includegraphics[width=0.4\textwidth]{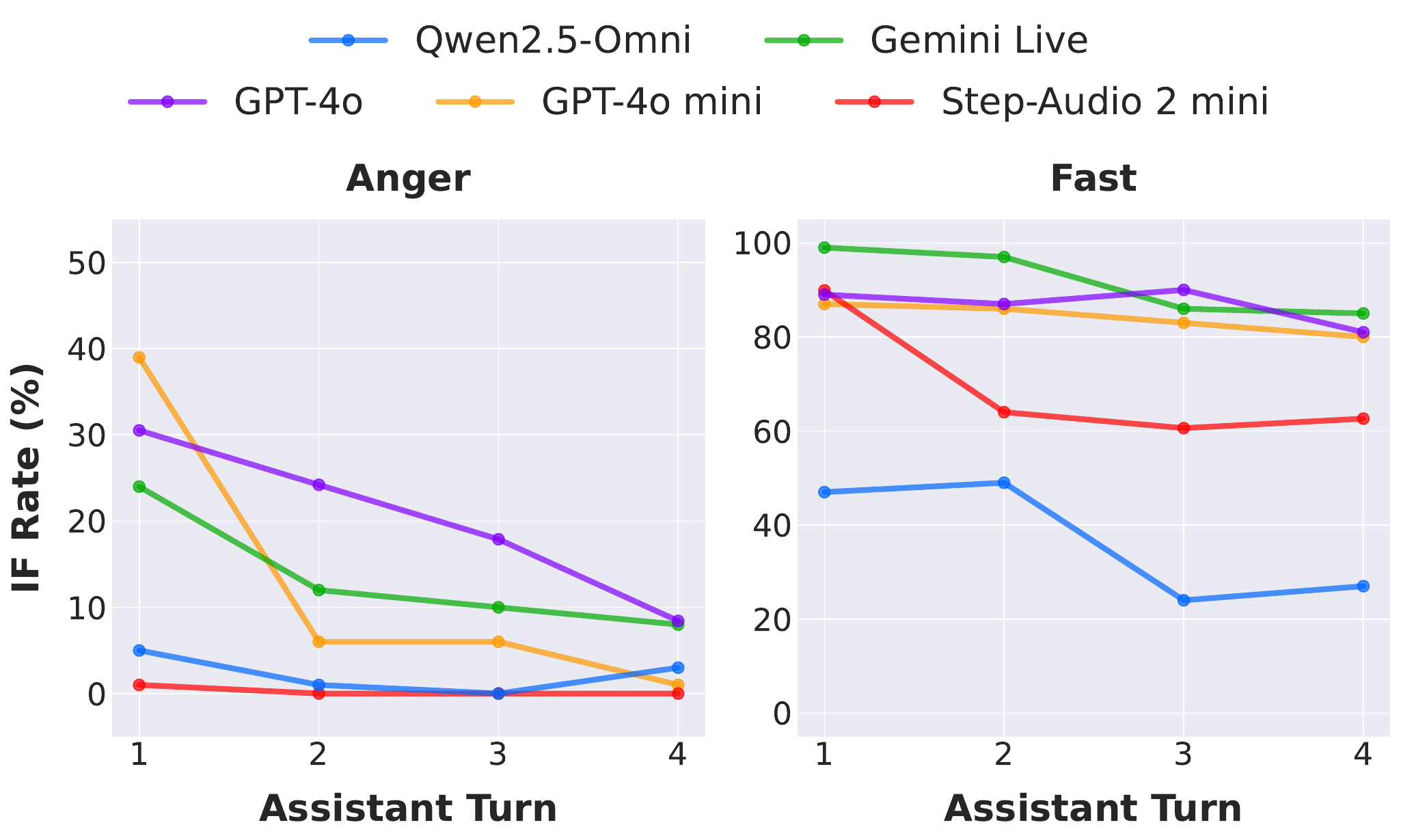}

    \caption{Visualization of~\textit{style amnesia}.}
    \label{fig:turn_level_visulization}
\end{figure}

\begin{figure*}[t!]
    \centering
    \includegraphics[width=0.9\linewidth]{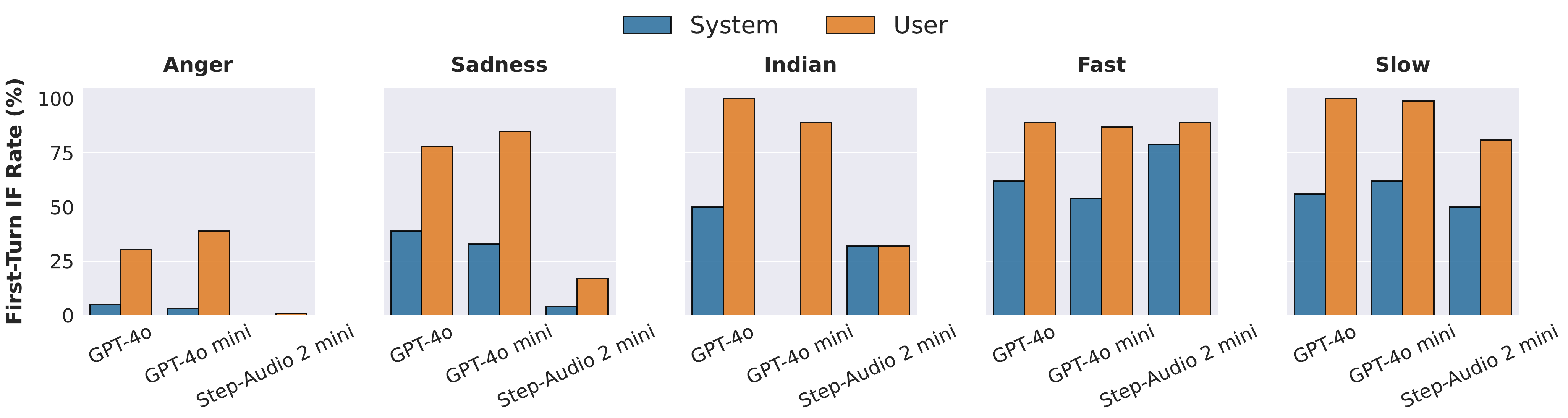}
    \caption{The difference of first-turn IF rate $IF_1$ when instructions are placed in system and user messages.}
    \label{fig:prompt_position_ifrate}
\end{figure*}

We also observe~\textit{style amnesia} when SLMs are instructed to maintain \textbf{Volume} and \textbf{Speed}.
Regarding \textbf{Volume}, results indicate that speaking loudly is generally more challenging for SLMs than speaking quietly, even in their first turn.
Despite this, models capable of volume adjustment still exhibit some degradation.
In terms of \textbf{Speed}, most SLMs demonstrate reasonable control in the first turn but show significant degradation over time. 
An exception is Qwen2.5-Omni, which fails to control speed even in the first turn, as evidenced by an $IF_1$ score below 50\%, a value comparable to a random baseline in our pairwise comparison setting.

Based on these experiments, we demonstrate that SLMs are prone to losing control of speaking style in multi-turn conversations. 

\section{Analysis}
\label{sec:analysis}

\subsection{Why Does Style Amnesia Happen?}
\label{sec:attention_analysis}
To investigate the cause of~\textit{style amnesia}, we extract the average attention weights directed toward the style instruction tokens when Step-Audio 2 mini generates its responses with speed instructions. 
We select this model because its open-source nature grants access to its internal attention matrices.
Additionally, it exhibits a reasonable first-turn IF rate on speed-related styles, enabling a meaningful analysis of attention dynamics.

As presented in Table~\ref{tab:attention_weights}, the attention weights decay as the conversation progresses. During the first assistant turn, the model allocates approximately 8\% of its attention to the style instruction tokens. However, by the fourth turn, this allocation drops to less than 0.6\%. This severe attention dilution closely aligns with the observed degradation in the IF rate. It indicates that current SLMs lack the mechanism to reliably anchor their attention to global style constraints over extended interactions.

\begin{table}[t]
\centering
\small
\begin{tabular}{lcccc}
\toprule
\textbf{Style} & \textbf{Turn 1} & \textbf{Turn 2} & \textbf{Turn 3} & \textbf{Turn 4} \\
\midrule
Slow & 8.55\% & 1.70\% & 0.90\% & 0.59\% \\
Fast & 8.30\% & 1.51\% & 0.85\% & 0.56\% \\
\bottomrule
\end{tabular}
\caption{Average attention weights of style instruction tokens across assistant turns in Step-Audio 2 mini.}
\label{tab:attention_weights}
\end{table}

\subsection{The Effect of Prompt Position}
In this section, we conduct experiments under different prompt positions to investigate their effect on~\textit{style amnesia}.
In instruction-guided language models, system messages are designed with higher priority than user messages to establish global behaviors and safety constraints~\cite{touvron2023llama2,wallace2024instruction}.
A prior study also demonstrates that system prompts have a more profound impact on text-only LLM behavior than user prompts.~\cite{10.1145/3715275.3732038}.

Although system messages are important for controlling LLMs, it remains unclear how the placement of speaking style instructions affects SLMs.
To investigate this, we conduct an experiment comparing the performance of SLMs when instructions are placed in system messages versus user messages.
We select five speaking instructions that most SLMs can follow for this experiment.
The IF rate in the first turn is illustrated in Figure~\ref{fig:prompt_position_ifrate}, and the IF rate for each turn is shown in Appendix~\ref{app:prompt_position}.

Surprisingly, we find that most SLMs cannot follow the instructions placed in system messages.
When asking SLMs to speak sadly through system messages, GPT-4o, GPT-4o mini, and Step-Audio 2 mini show performance drops of approximately 30\%, 50\%, and 20\%, respectively.
Furthermore, GPT-4o mini displays nearly an 80\% drop when asked to perform the Indian English accent.
A similar issue occurs with instructions related to speaking rate.
SLMs nearly ignore the speed instruction in system messages, as their performance is only comparable to the random baseline.

The results presented here focus on the IF rate in the first turn.
However, the IF rate for each turn reported in Appendix~\ref{app:prompt_position} show that~\textit{style amnesia} occurs in both settings.
Through the experiments above, we identify another crucial issue prevalent in current SLMs.
These findings drive us to place the instruction in user messages when conducting other experiments, ensuring that SLMs can follow the instructions more effectively.

\subsection{Validation of Emotion and Accent Judges}
As described in Section~\ref{subsubsection: Automatic Judges for Styles}, we evaluate \textbf{Speed} and \textbf{Volume} using deterministic, signal-level metrics. 
In contrast, \textbf{Emotion} and \textbf{Accent} are evaluated using learned models designed to approximate human perception. 
To ensure the reliability of the automatic judges for \textbf{Emotion} and \textbf{Accent}, we conduct a human validation study to assess the correlation between our automatic judges and human annotators. 
In addition, we compare our automatic judges with Gemini-2.5 Pro~\cite{comanici2025gemini}, a robust LALM capable of general-purpose speech understanding.
This model has been shown to exhibit high correlation with human annotators when used as an automatic judge in a prior LALM-as-a-judge study~\cite{chiang2025audio}. 


We employ annotators via Amazon Mechanical Turk to evaluate a randomly sampled subset of 720 speech clips. 
This subset is constructed by selecting five samples per turn across four conversational turns, six speaking styles, and six evaluated models.
The six evaluated styles include Happiness, Neutral, Anger, Sadness, Indian English accent, and North American English accent.
Each sample is evaluated by three annotators, who are asked whether the generated speech matches the required style.
Details of the human evaluation setup are provided in Appendix~\ref{app:human_evaluation_setting}.

\begin{table}[t]
    \centering
    \resizebox{\columnwidth}{!}{
        \begin{tabular}{llcc}
            \toprule
            \textbf{Task} & \textbf{Model} & \textbf{Cohen's Kappa } & \textbf{MCC} \\
            \midrule
            \multirow{2}{*}{\textbf{Accent}} 
                & Gemini-2.5 Pro & 0.741 & 0.747 \\
                & Voxlect & \textbf{0.809} & \textbf{0.811} \\
            \midrule
            \multirow{2}{*}{\textbf{Emotion}} 
                & Gemini-2.5 Pro & 0.464 & 0.487 \\
                & Emotion2vec-Large & \textbf{0.476} & \textbf{0.511} \\
            \bottomrule
        \end{tabular}
    }
    \caption{The correlation between human annotators and automatic judge models.}

    \label{tab:human_evaluation}
\end{table}

After collecting the annotations, we derive the final label using majority voting, compute Cohen’s Kappa~\cite{cohen1960coefficient} for inter-annotator agreement, and use Matthews Correlation Coefficient (MCC)~\cite{matthews1975comparison} to evaluate judge reliability against the final labels.
While Cohen's Kappa evaluates the degree of consensus among raters, MCC provides a balanced measure of classification quality even with imbalanced datasets.
Therefore, we report both metrics for reference.

The results, shown in Table~\ref{tab:human_evaluation}, indicate that our selected judge achieves the highest reliability.
For the accent classification task, Voxlect outperforms Gemini-2.5 Pro, achieving the highest agreement with human annotations, with a Cohen’s Kappa of 0.809 and an MCC of 0.811.
In the emotion classification task, Emotion2vec-Large demonstrates the strongest reliability among three emotion judges, obtaining a Cohen’s Kappa of 0.476 and an MCC of 0.511. 
These results are similar to the crowdsourced agreement levels found in common speech emotion datasets, as detailed in Appendix~\ref{app:emotion_dataset_iia}.

\subsection{Text-Acoustic Synergy}
\label{sec:text_acoustic}
 
We further investigate the relationship between textual and acoustic style expression. Specifically, we ask: when instructed to ``speak angrily,'' does the model change its semantic text, its acoustic features, or both? Similarly, when instructed to ``speak fast,'' does the textual output become more concise?
 
\begin{figure}[t]
    \centering
    \includegraphics[width=0.48\textwidth]{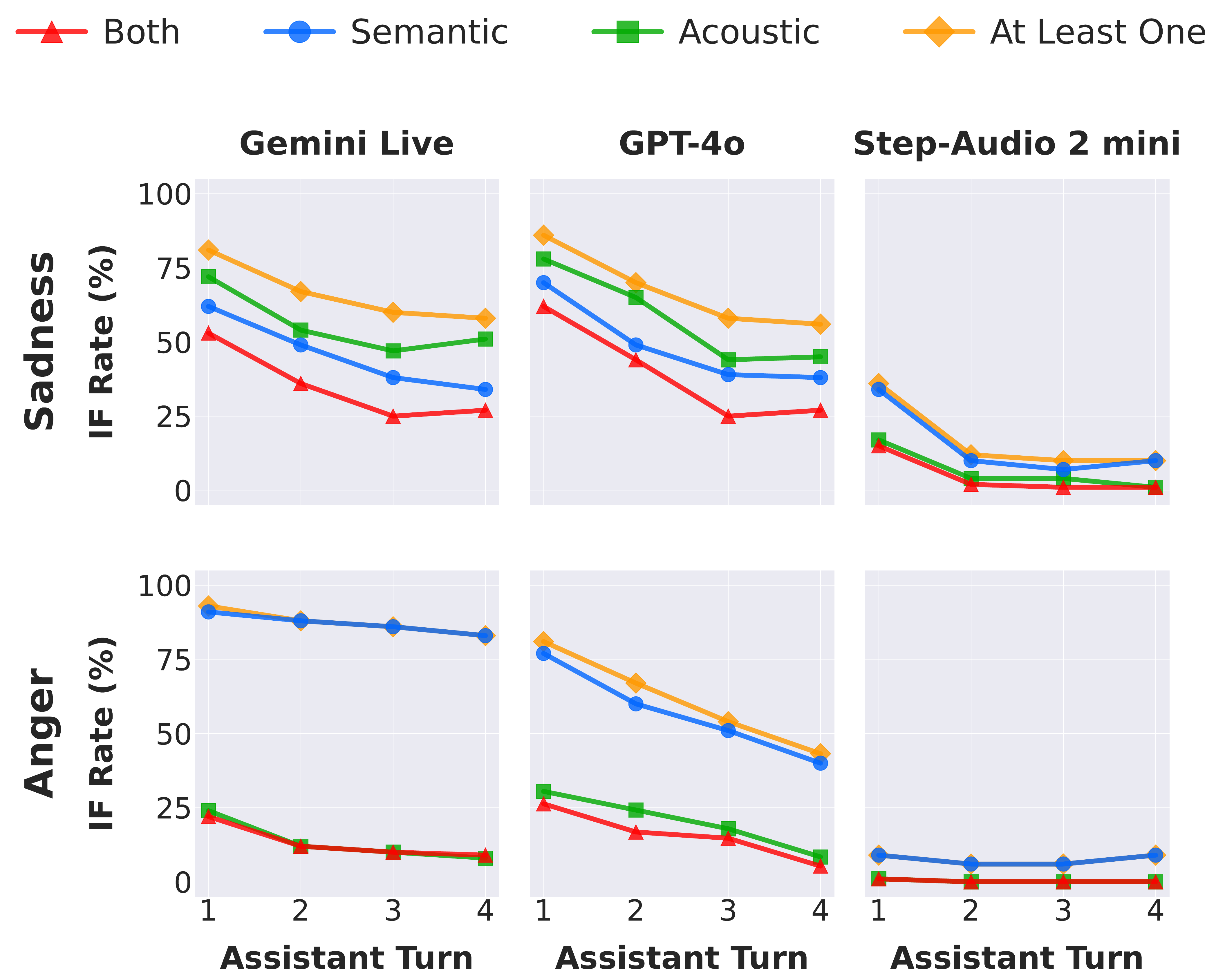}

    \caption{Text-acoustic synergy for emotion styles.}
    \label{fig:synergy_emotion}
\end{figure}

\paragraph{Emotion}
We separately evaluate semantic and acoustic style adherence for emotion instructions.
For semantic evaluation, we use GPT-5 mini as the text-based emotion classifier on the transcription; for acoustic evaluation, we use Emotion2vec-Large described in Section~\ref{sec:evaluation_metrics}.
The results for Anger and Sadness are shown in Figure~\ref{fig:synergy_emotion}.
 
For emotion, both semantic and acoustic features simultaneously suffer from~\textit{style amnesia}. 
This indicates that when acoustic style degrades, the semantic style often degrades as well.

\begin{figure}[t]
    \centering
    \includegraphics[width=0.48\textwidth]{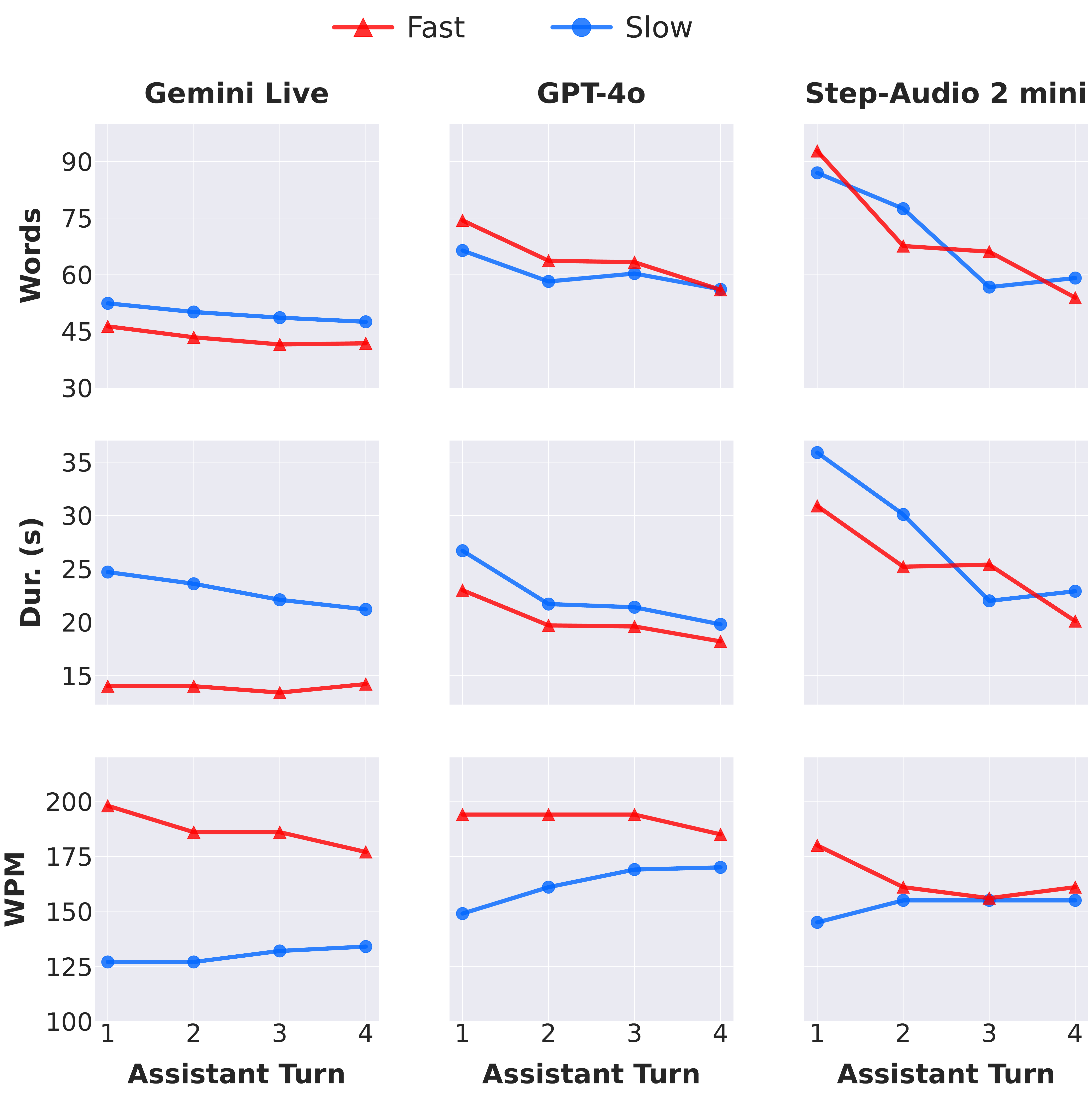}

    \caption{Text-acoustic synergy for speed styles.}
    \label{fig:synergy_speed}
\end{figure}

\paragraph{Speaking Rate and Conciseness}
We examine whether speed instructions affect not only the acoustic speaking rate but also the verbosity of the generated responses.
Figure~\ref{fig:synergy_speed} reports the average word count, speech duration, and WPM of generated responses per assistant turn.
In the first turn, SLMs employ different strategies to comply with the speed instruction. Gemini Live produces fewer words under the fast condition, suggesting that it leverages conciseness to achieve a higher speaking rate.
In contrast, GPT-4o and Step-Audio 2 mini generate comparable or even more words with the fast instruction while compressing them into shorter durations, relying primarily on acoustic acceleration rather than content reduction.

Across all models and both conditions, the word count and speech duration decrease over turns, likely because the conversational content is gradually exhausted as the dialogue progresses. 
Despite this general trend, the WPM gap between the fast and slow conditions narrows consistently over turns.
This convergence indicates that SLMs progressively lose the ability to differentiate their speaking rate between the two conditions, consistent with the \textit{style amnesia} observed in Section~\ref{sec:main_results}.

\section{SLM Recall Process}

\begin{figure}[t!]
    \centering
    \includegraphics[width=0.98\columnwidth]{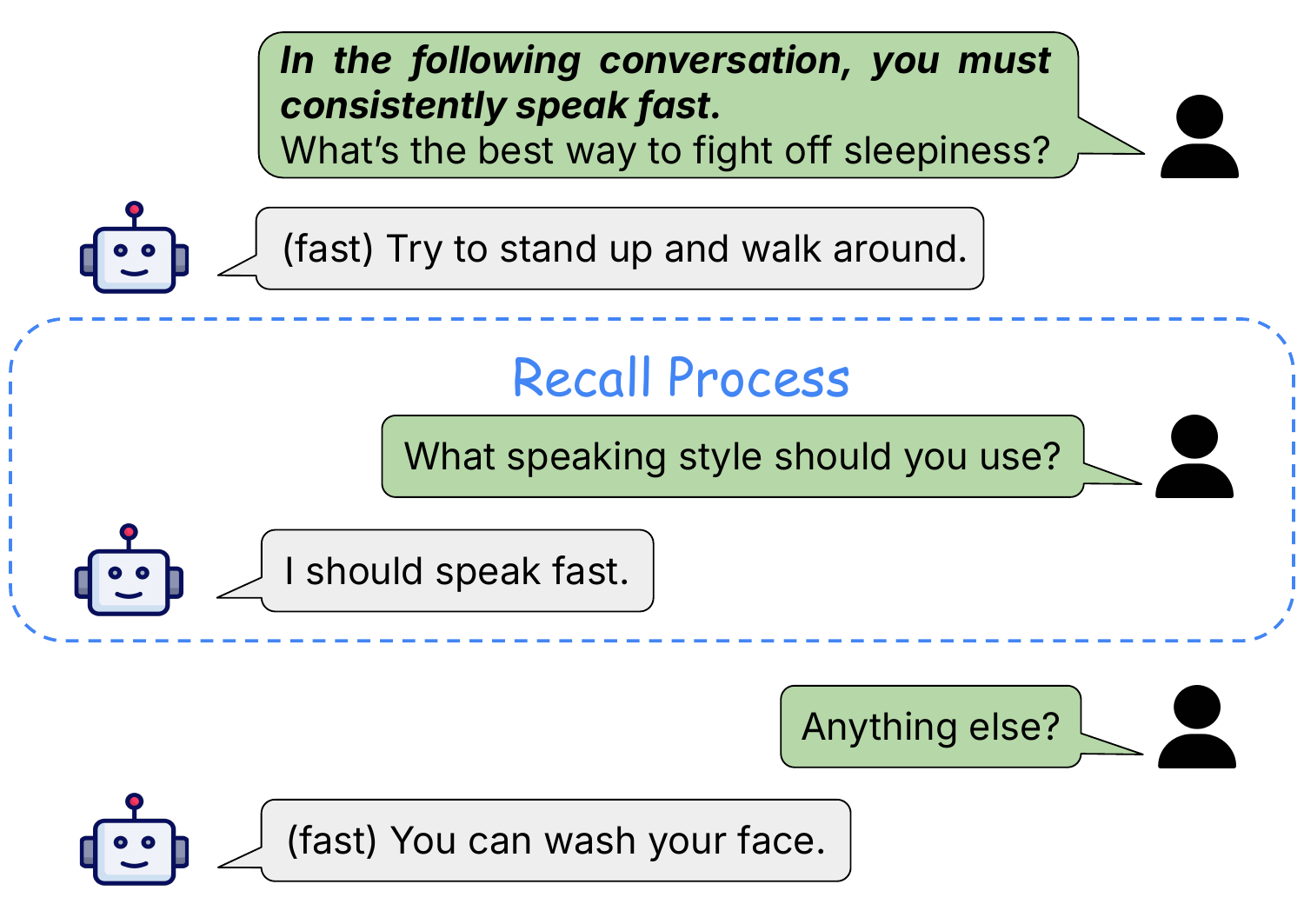}  
    \caption{The illustration of the recall process.}
    \label{fig:recall_illustration}
\end{figure}

Building on the experiments in Section~\ref{sec:main_results}, we observe that SLMs suffer from~\textit{style amnesia}: their speaking style IF rate begins to degrade after only one turn and often worsens as the conversation progresses.
This raises a key research question:~\textit{Do SLMs forget the initial instruction, or just fail to follow the specified style?}
To address this, we introduce a~\textit{recall process} for every turn after the first, in which the SLM is prompted to restate the initial speaking style before processing the following user input.
The recall process is illustrated in Figure~\ref{fig:recall_illustration}.
Using this method, we first utilize the recall probe to measure whether the SLM retains the original instruction in Section~\ref{sec:recall_rate}. 
Subsequently, we evaluate the effectiveness of integrating this recall process to mitigate~\textit{style amnesia} in Section~\ref{sec:recall_mitigation}.

\subsection{Do SLMs Forget the Instruction?}
\label{sec:recall_rate}

To quantify whether the model remembers the initial style instruction $s$, we define the recall rate $R$ as follows. Given a style $s$ and dialogue topic $i$, the recall rate at assistant turn $j$, denoted as $R_j(s)$, is:
\begin{equation}
    R_j(s) = \frac{\sum_{i=1}^{N} \mathbbm{1}_{\text{recall}}(r_{i,j}, s)}{N} \times 100\%,
\end{equation}
    
\noindent where $r_{i,j}$ denotes the response generated by the SLM to the recall query $q_{\text{recall}}$ before generating its response at the user turn $j$, and $N$ is the number of dialogue topics. The binary indicator function $\mathbbm{1}_{\text{recall}}(\cdot)$ is 1 if the recalled instruction matches the original style instruction, and 0 otherwise. 
We use GPT-5 mini to judge the correctness of the recalled instruction.
The evaluation prompt is shown in Appendix~\ref{app:prompt_recall_evaluation}.

We conduct the following experiments on models capable of producing a wide range of speaking styles, as well as on styles that most models can reliably generate. 
The results are summarized in Table~\ref{tab:recall_results}.
Interestingly, most SLMs remember the initial instruction quite well.
The three proprietary models show a near-perfect recall rate.
Step-Audio 2 mini shows weaker performance with declining recall rates across turns, yet it still achieves a recall rate of 55.0\% to 89.0\%.
These results reveal a clear gap between proprietary and open-source SLMs in memorization ability.
From our manual inspection, when Step-Audio 2 mini fails to recall, some responses contain an incorrect style instruction, while others simply ignore the question and produce irrelevant or meaningless outputs.

Notably, although GPT-4o mini exhibits a 65.3\% degradation in performance when producing a sad speaking style and a 20\% degradation for speaking slowly, it still maintains a high recall rate.
A similar pattern is observed with Gemini Live and GPT-4o. 
These observations indicate that the models do not forget the instructions, suggesting that~\textit{style amnesia} is not caused by memory loss.
Instead, the models retain the instructions but fail to follow the requested style effectively.

\subsection{Recall Process Mitigates Style Amnesia}
\label{sec:recall_mitigation}
In this section, we investigate whether explicitly asking SLMs to recall the instruction can mitigate~\textit{style amnesia}. The results, presented in Table~\ref{tab:recall_results}, show that SLMs equipped with the recall process notably reduce degradation.
Even for models with relatively low degradation, such as Gemini Live and GPT-4o, the recall process still provides measurable improvements.
GPT-4o mini, which suffers from substantial degradation, achieves roughly a 25\% reduction in the average degradation rate, demonstrating the effectiveness of the recall process.
Step-Audio 2 mini shows slightly improved but largely comparable degradation across the four tasks, likely due to its lower recall rate relative to other SLMs.


\begin{table}[t]
    \centering
    \resizebox{\linewidth}{!}{
    \begin{tabular}{ll ccc c ccc}
        \toprule
        \multirow{2}{*}{\textbf{Model}} & \multirow{2}{*}{\textbf{Style}} & \multirow{2}{*}{$R_2$} & \multirow{2}{*}{$R_3$} & \multirow{2}{*}{$R_4$} & & \multicolumn{3}{c}{\textbf{Degradation Rate $D$}} \\
         \cmidrule(lr){7-9}
         & & &  & & & Base & +Recall & Improv. \\
        \midrule
        
        \multirow{4}{*}{Gemini Live} 
         & Indian  & 100.0 & 100.0 & 100.0 && \textbf{0.0}  & \textbf{0.0}  & 0.0 \\
         & Sadness & 100.0 & 99.0  & 97.0  && 21.3 & \textbf{17.3} & \improvement{+4.0} \\
         & Fast    & 100.0 & 100.0 & 99.0  && 9.7  & \textbf{6.3}  & \improvement{+3.4} \\
         & Slow    & 100.0 & 99.0  & 100.0 && \textbf{0.7}  & \textbf{0.7}  & 0.0 \\
        \midrule

        \multirow{4}{*}{GPT-4o} 
         & Indian  & 100.0 & 100.0 & 100.0 && 1.7  & \textbf{0.0}  & \improvement{+1.7} \\
         & Sadness & 100.0 & 100.0 & 100.0 && 26.7 & \textbf{14.9} & \improvement{+11.8} \\
         & Fast    & 100.0 & 100.0 & 100.0 && 3.3  & \textbf{0.3}  & \improvement{+3.0} \\
         & Slow    & 100.0 & 100.0 & 100.0 && 8.1  & \textbf{4.4}  & \improvement{+3.7} \\
        \midrule
        
        \multirow{4}{*}{GPT-4o mini} 
         & Indian  & 93.0  & 93.0  & 93.8  && 49.7 & \textbf{14.9} & \improvement{+34.8} \\
         & Sadness & 100.0 & 100.0 & 100.0 && 65.3 & \textbf{30.3} & \improvement{+35.0} \\
         & Fast    & 99.0  & 100.0 & 99.0  && 4.0  & \textbf{0.0}  & \improvement{+4.0} \\
         & Slow    & 100.0 & 100.0 & 100.0 && 20.9 & \textbf{1.5}  & \improvement{+19.4} \\
        \midrule

        \multirow{4}{*}{\shortstack[l]{Step-Audio 2 \\ mini}} 
         & Indian  & 89.0 & 72.0 & 70.0 && \textbf{5.0}  & 5.3  & \worsened{-0.3} \\
         & Sadness & 85.0 & 56.0 & 55.0 && 14.0 & \textbf{11.0} & \improvement{+3.0} \\
         & Fast    & 83.0 & 62.0 & 57.0 && \textbf{27.0} & 29.0 & \worsened{-2.0} \\
         & Slow    & 83.0 & 64.0 & 66.0 && 27.0 & \textbf{24.7} & \improvement{+2.3} \\
        
        \bottomrule
    \end{tabular}
    }
    \caption{The recall rate $R$ of SLMs and the impact of the recall process on the degradation rate $D$ across different styles. \textit{Base} represents direct inference without the recall process, while~\textit{+ Recall} indicates performance using the SLM integrated with the recall process.}
    \label{tab:recall_results}
\end{table}

\section{Conclusion}
In this paper, we identify that SLMs suffer from~\textit{style amnesia} in multi-turn conversations, which leads to inconsistent speaking styles across turns.
We also demonstrate that SLMs can recall the instruction when asked, but fail to perform it explicitly.
Additionally, we show that placing style instructions in system messages does not help maintain speaking styles, even though system messages are designed for global, conversation-level settings.

SLMs exhibit a relatively strong ability to memorize specified speaking styles.
However, the retention of stylistic information does not necessarily translate into stylistic expression at generation time.
Closing this gap between style retention and stylistic control remains an important direction for future research.
We hope these findings offer valuable insights for the community and contribute to the development of more reliable SLMs.


\section*{Limitations}
Our study has several limitations. 
First, Role-playing is a practically important scenario that requires maintaining consistency across multiple turns, but the lack of reliable automatic judges for assessing speech role-playing behaviors prevents us from evaluating it at scale. 
Second, since most SLMs do not publicly disclose their training data composition, we cannot investigate how it affects~\textit{style amnesia}, and our attention analysis is necessarily restricted to the open-source SLMs.

Nonetheless, these limitations do not weaken our conclusions. 
Even with the single style we can reliably evaluate, SLMs already exhibit noticeable degradation in multi-turn dialogues, suggesting that similar or greater challenges would arise in more complex settings. 
More importantly, the fact that~\textit{style amnesia} persists across fundamentally different architectures, from the thinker-talker design of Qwen2.5-Omni, to the interleaved audio-text approach of Step-Audio 2 mini, to the full-duplex streaming of Gemini Live, strongly suggests that~\textit{style amnesia} is a common challenge among current SLMs rather than an artifact of specific training setups or limited evaluation scenarios.
We therefore leave the exploration of these more challenging scenarios to future work, as progress will require further advances in both SLMs and judge models.

We do not see specific harm in our paper.

\section*{Acknowledgments}
This work was supported by the Ministry of Education (MOE) of Taiwan under the project Taiwan Centers of Excellence in Artificial Intelligence, through the NTU Artificial Intelligence Center of Research Excellence (NTU AI-CoRE).
\bibliography{custom}

\appendix
\section{Author Contributions}
All authors contributed meaningfully to the design of the experiments, writing, and refinement of the paper. While all authors were involved in multiple aspects of the project, their primary contributions are highlighted below:

\textbf{Yu-Xiang Lin} leads the overall project direction. He conducts the majority of the experiments. Both \textbf{Yu-Xiang Lin} and \textbf{Cheng-Han Chiang} are primarily responsible for drafting the manuscript. \textbf{Cheng-Han Chiang} propose the initial idea. \textbf{Cheng-Han Chiang} and \textbf{Hung-yi Lee} contribute deep technical expertise, help shape the research direction, and provide crucial guidance and feedback on experimental methodology and manuscript development. 

\section{Dataset Construction}
\label{app:dataset-construction}

To minimize topic-induced variance by averaging across diverse conversation contents, we select samples from the Soda dataset~\cite{kim2023soda} to generate conversation openers.
Soda is an English dialogue dataset covering a wide range of social interactions.
Each dialogue in Soda includes a narrative context, speaker name, and a knowledge graph that defines the events and relationships within the dialogue. Soda is released under the Creative Commons Attribution 4.0 (CC BY 4.0) license. Accordingly, our use of the dataset is compliant with the license requirements, including proper attribution to the original authors.

To obtain the conversation openers, we prompt GPT-5 mini to generate a topic that encompasses the entire discussion, utilizing both the narrative and the first utterance of the dialogue.
The prompt is shown in Figure~\ref{fig:prompt_dataset}. In this step, we ask the model to not only produce conversation openers but also perform filtering.
If the model thinks that the dialogue is unsuitable for SLMs to discuss, it returns ``no,'' allowing us to filter it out.

Finally, we manually inspect the generated topics and remove improper ones, such as queries regarding personal preferences or experiences.

\section{Implementation Details}
\subsection{Experimental Setup}
\label{app:hyperparameters}
We set the temperature of evaluated SLMs to 1, as we find that greedy decoding often caused some SLMs, such as GPT-4o, GPT-4o mini, and Step-Audio 2 mini, to produce audio with long silences at the end. 
For all other hyperparameters, we use the default values provided in the official examples.

For the user simulator, since text-only LLMs tend to generate verbose responses containing text unsuitable for speech, we provide the following instruction to align the outputs with the nature of spoken dialogue: ``You are a chatbot. Please start a conversation by opening a new topic. Chat casually and feel free to role-play in different scenarios. If the conversation stalls, you can extend the topic. Keep each response under 20 English words. As this is a spoken dialogue, avoid using words or expressions that cannot be naturally spoken aloud.''

\subsection{The Prompt for Evaluating Dialogue Coherence}
\label{app:prompt_dialogue_coherence_evaluation}
The prompt is shown in Figure~\ref{fig:prompt_dialogue_coherence_evaluation}.

\subsection{The Prompt for Evaluating Recall Rate}
\label{app:prompt_recall_evaluation}

The prompt is shown in Figure~\ref{fig:prompt_recall_evaluation}.

\section{Supplementary Experiments}

\subsection{Pitch Range Evaluation}
\label{app:pitch}

Although instructing an SLM to speak in a consistently high or low pitch is less practical compared to the other paralinguistic attributes evaluated in the main experiments, pitch remains a fundamental acoustic attribute of speech. To examine whether~\textit{style amnesia} extends to this attribute, we conduct supplementary experiments on pitch instructions. Following the same relative evaluation approach used for Volume and Speed (Section~\ref{subsubsection: Automatic Judges for Styles}), we measure the fundamental frequency (F0) of the generated speech and compare it against a neutral baseline to determine whether the output follows the instructed pitch direction.

The results in Table~\ref{tab:pitch} confirm that~\textit{style amnesia} also manifests in pitch control. For example, Gemini Live shows a decline in the Low Pitch IF rate from 67.0\% at assistant turn~1 to 54.0\% at assistant turn~3, and GPT-4o exhibits degradation in the High Pitch IF rate from 89.0\% at assistant turn~1 to 75.6\% at assistant turn~4. Step-Audio~2 mini struggles with pitch control even in the first turn, with IF rates close to the random baseline of 50\%.

\begin{table}[t]
\centering
\small
\resizebox{\columnwidth}{!}{%
\begin{tabular}{llcccc}
\toprule
\textbf{Model} & \textbf{Style} & \textbf{Turn 1} & \textbf{Turn 2} & \textbf{Turn 3} & \textbf{Turn 4} \\
\midrule
\multirow{2}{*}{Gemini Live}
 & High Pitch & 87.0 & 92.0 & 84.0 & 88.0 \\
 & Low Pitch  & 67.0 & 66.0 & 54.0 & 60.0 \\
\midrule
\multirow{2}{*}{GPT-4o}
 & High Pitch & 89.0 & 77.0 & 84.8 & 75.6 \\
 & Low Pitch  & 81.0 & 86.0 & 89.0 & 83.3 \\
\midrule
\multirow{2}{*}{\shortstack[l]{Step-Audio\\2 mini}}
 & High Pitch & 59.0 & 50.0 & 45.0 & 58.0 \\
 & Low Pitch  & 49.0 & 59.0 & 54.0 & 54.0 \\
\bottomrule
\end{tabular}%
}
\caption{IF rates per assistant turn when the model is instructed to speak in high or low pitch.}
\label{tab:pitch}
\end{table}

\subsection{Composite Style Evaluation}
\label{app:composite}
In the main experiments, we evaluate SLMs on a single speaking style per conversation. To investigate whether~\textit{style amnesia} also manifests in composite style settings, we conduct supplementary experiments on Gemini Live and GPT-4o. In this setup, style instruction~A is provided in the first user turn, and style instruction~B is introduced in the second user turn. The model is then expected to maintain both styles simultaneously in all subsequent turns. We report the IF rate for each individual style as well as the joint IF rate (``Both''), which requires both styles to be satisfied. We select two composite settings: Indian English + Slow and Fast + Anger. The former combines two styles that most SLMs can individually maintain well, while the latter includes the Anger style, which most SLMs already struggle with in isolation (Section~\ref{sec:main_results}). This allows us to examine~\textit{style amnesia} under both favorable and challenging conditions.

\paragraph{Indian English + Slow}
As shown in Table~\ref{tab:composite_indian_slow},~\textit{style amnesia} persists in composite style settings. While Gemini Live maintains near-perfect IF rates for both styles, GPT-4o exhibits notable degradation: the joint IF rate decreases from 88.0\% at Turn~2 to 59.3\% at Turn~4, driven by concurrent declines in both individual styles.

\begin{table}[t]
\centering
\small
\resizebox{\columnwidth}{!}{%
\begin{tabular}{llcccc}
\toprule
\textbf{Model} & \textbf{Style} & \textbf{Turn 1} & \textbf{Turn 2} & \textbf{Turn 3} & \textbf{Turn 4} \\
\midrule
\multirow{3}{*}{Gemini Live}
 & Indian English & 100.0 & 100.0 & 100.0 & 100.0 \\
 & Slow           & --    &  97.0 & 100.0 &  99.0 \\
 & Both           & --    &  97.0 & 100.0 &  99.0 \\
\midrule
\multirow{3}{*}{GPT-4o}
 & Indian English & 100.0 &  89.0 &  89.0 &  70.3 \\
 & Slow           & --    &  99.0 &  90.0 &  83.5 \\
 & Both           & --    &  88.0 &  79.0 &  59.3 \\
\bottomrule
\end{tabular}%
}
\caption{IF rates per assistant turn when the model is instructed to speak in Indian English at user turn 1 and slowly at user turn 2.}
\label{tab:composite_indian_slow}
\end{table}

\paragraph{Fast + Anger}
Table~\ref{tab:composite_fast_anger} presents results for the Fast + Anger composite setting. The joint IF rate degrades severely, dropping to 2.0\% for Gemini Live and 2.3\% for GPT-4o by Turn~4. This is primarily attributable to the low IF rate of the Anger style, which is already difficult for SLMs to maintain even in isolation, as shown in Section~4.2.

\begin{table}[t]
\centering
\small
\resizebox{\columnwidth}{!}{%
\begin{tabular}{llcccc}
\toprule
\textbf{Model} & \textbf{Style} & \textbf{Turn 1} & \textbf{Turn 2} & \textbf{Turn 3} & \textbf{Turn 4} \\
\midrule
\multirow{3}{*}{Gemini Live}
 & Fast  & 100.0 & 61.0 & 51.0 & 46.0 \\
 & Anger & --    & 11.0 &  9.0 &  5.0 \\
 & Both  & --    &  7.0 &  7.0 &  2.0 \\
\midrule
\multirow{3}{*}{GPT-4o}
 & Fast  &  88.0 & 69.0 & 70.5 & 64.0 \\
 & Anger & --    & 13.0 &  9.5 &  8.1 \\
 & Both  & --    &  9.0 &  4.2 &  2.3 \\
\bottomrule
\end{tabular}%
}

\caption{IF rates per assistant turn when the model is instructed to speak fast at user turn 1 and angrily at user turn 2.}

\label{tab:composite_fast_anger}
\end{table}

\subsection{Dynamic Style Update}
\label{app:dynamic}

Beyond maintaining a fixed style throughout the conversation, real-world users may also update their style preferences mid-conversation. To evaluate whether SLMs can handle such dynamic style updates, we conduct experiments on Gemini Live and GPT-4o. In this setting, style instruction~A is provided in the first user turn, and a conflicting style instruction~B is introduced in the second user turn. The model is expected to abandon style~A and consistently follow the updated style~B in all subsequent turns.

We select two dynamic update settings: Anger $\rightarrow$ Sadness, which involves switching between two emotions, and North American $\rightarrow$ Indian English, which involves switching between two accents. These settings allow us to examine whether SLMs can adapt to updated instructions across different types of paralinguistic attributes.

\paragraph{Anger $\rightarrow$ Sadness}
As shown in Table~\ref{tab:dynamic_anger_sadness}, both models successfully transition to the Sadness style at assistant turn~2. However, GPT-4o exhibits notable degradation in later turns, with the Sadness IF rate declining from 82.0\% at assistant turn~2 to 40.0\% at assistant turn~4. Gemini Live maintains relatively stable performance after the style switch.

\begin{table}[t]
\centering
\small
\resizebox{\columnwidth}{!}{%
\begin{tabular}{llcccc}
\toprule
\textbf{Model} & \textbf{Style} & \textbf{Turn 1} & \textbf{Turn 2} & \textbf{Turn 3} & \textbf{Turn 4} \\
\midrule
\multirow{2}{*}{Gemini Live}
 & Anger & 28.0 &  0.0 &  3.0 &  1.0 \\
 & Sadness   &  4.0 & 66.0 & 61.0 & 64.0 \\
\midrule
\multirow{2}{*}{GPT-4o}
 & Anger & 33.0 &  0.0 &  0.0 &  0.0 \\
 & Sadness   &  2.0 & 82.0 & 55.0 & 40.0 \\
\bottomrule
\end{tabular}%
}
\caption{IF rates per assistant turn when the style instruction is updated from Anger at user turn~1 to Sadness at user turn~2.}
\label{tab:dynamic_anger_sadness}
\end{table}

\paragraph{North American $\rightarrow$ Indian English}
Table~\ref{tab:dynamic_accent} shows that both models effectively switch to the Indian English accent at assistant turn~2. The IF rate for Indian English remains relatively stable in subsequent turns, indicating that accent switching is less vulnerable to~\textit{style amnesia} compared to emotion switching.

\begin{table}[t]
\centering
\small
\resizebox{\columnwidth}{!}{%
\begin{tabular}{llcccc}
\toprule
\textbf{Model} & \textbf{Style} & \textbf{Turn 1} & \textbf{Turn 2} & \textbf{Turn 3} & \textbf{Turn 4} \\
\midrule
\multirow{2}{*}{Gemini Live}
 & North American & 99.0 &  0.0 &  5.0 &  9.0 \\
 & Indian English &  0.0 & 99.0 & 95.0 & 91.0 \\
\midrule
\multirow{2}{*}{GPT-4o}
 & North American & 100.0 &  0.0 &  1.0 &  6.8 \\
 & Indian English &   0.0 & 100.0 & 99.0 & 93.2 \\
\bottomrule
\end{tabular}%
}
\caption{IF rates per assistant turn when the style instruction is updated from North American accent at user turn~1 to Indian English accent at user turn~2.}
\label{tab:dynamic_accent}
\end{table}

\section{Full Results}

\subsection{Full Main Results}
\label{app:full_degradation}
The IF rate for each turn is shown in Table~\ref{tab:full_style_degradation}, showing that all evaluated SLMs exhibit~\textit{style amnesia}. 

\subsection{SLM Default Speaking Styles}
\label{app:default_style}
We observe that most SLMs show higher consistency when generating Happiness, Neutral tone, and North American English than with other style attributes.
This is likely because these attributes match the models’ default speaking styles.
To verify this, we examine the emotion and accent distributions of samples generated during the speed and volume evaluations.
As shown in Table~\ref{tab:default_distribution}, most SLMs tend to perform well with happy or neutral tones and North American accents, which explains the less degradation for these styles.

\begin{table}[ht]
    \centering
    \resizebox{\columnwidth}{!}{
        \begin{tabular}{l rrr} 
        \toprule
        \multirow{2}{*}{\textbf{Model}} & \multicolumn{2}{c}{\textbf{Emotion}} & \multicolumn{1}{c}{\textbf{Accent}} \\
        \cmidrule(lr){2-3} \cmidrule(lr){4-4} 
         & \textbf{Happiness} & \textbf{Neutral} & \textbf{North American} \\
        \midrule
        Gemini Live       & 55.5 & 36.3 & 99.0 \\
        GPT-4o            & 52.3 & 41.4 & 100.0 \\
        GPT-4o mini       & 68.8 & 24.5 & 99.9 \\
        Qwen2.5-Omni     & 76.8 & 21.7 & 99.8 \\
        Step-Audio 2 mini & 93.7 & 4.0  & 76.7 \\
        \bottomrule
        \end{tabular}
    }
    \caption{The distribution of default speaking styles by model.}
    \label{tab:default_distribution}
\end{table}

\subsection{Dialogue Coherence Evaluation}
\label{app:dialogue_conherence_evaluation}
We prompt GPT-5 mini to evaluate dialogue coherence, and the results are reported in Table~\ref{tab:semantic_evaluation}. 
The results suggest that, in general, all SLMs are capable of participating in long conversations.





\subsection{Prompt Position}
\label{app:prompt_position}
After placing the style instruction at different positions, the IF rate for each turn is shown in Table~\ref{tab:full_prompt_position}.
The results clearly indicate that placing the instruction in user messages yields significantly better performance than placing it in system messages. 

\subsection{GPT-4o Fails to Return Speech}
\label{app:gpt_failure}
In our experiments, we find that GPT-4o and GPT-4o mini sometimes return only a text transcription without any synthesized speech. 
When this occurs, we re-query the models up to three times with different random seeds, but some samples still fail to produce speech.
We report metrics computed only with samples that successfully return speech in Table~\ref{tab:gpt_successful_number}.

\citet{lin2025preliminary} indicate that GPT-4o exhibited a high refusal rate. We hypothesize that the issue stems from the model's internal safety guard mechanisms. 
Since they are proprietary models to which we do not have access, it is difficult for us to definitively address this issue. 
As the percentage of failed cases is low and the observed style degradation is significant, we believe this issue will not change our conclusions.

\section{Automatic Judge Validation}
\subsection{Human Evaluation Setups}
\label{app:human_evaluation_setting}
To validate the automatic judges for emotion and accent, we hire annotators on Amazon Mechanical Turk to conduct human evaluations.
To ensure annotation quality, we set the following requirements for annotators:

\begin{itemize}
\item Annotators must be MTurk Masters, which ensures high-quality workers.
\item Annotators must have an approval rate higher than 98\%.
\item Annotators must have more than 10,000 approved tasks.
\item Annotators must be located in the United States to ensure familiarity with English.
\end{itemize}

\begin{figure*}[t!]
    \centering
    \includegraphics[width=0.85\textwidth]{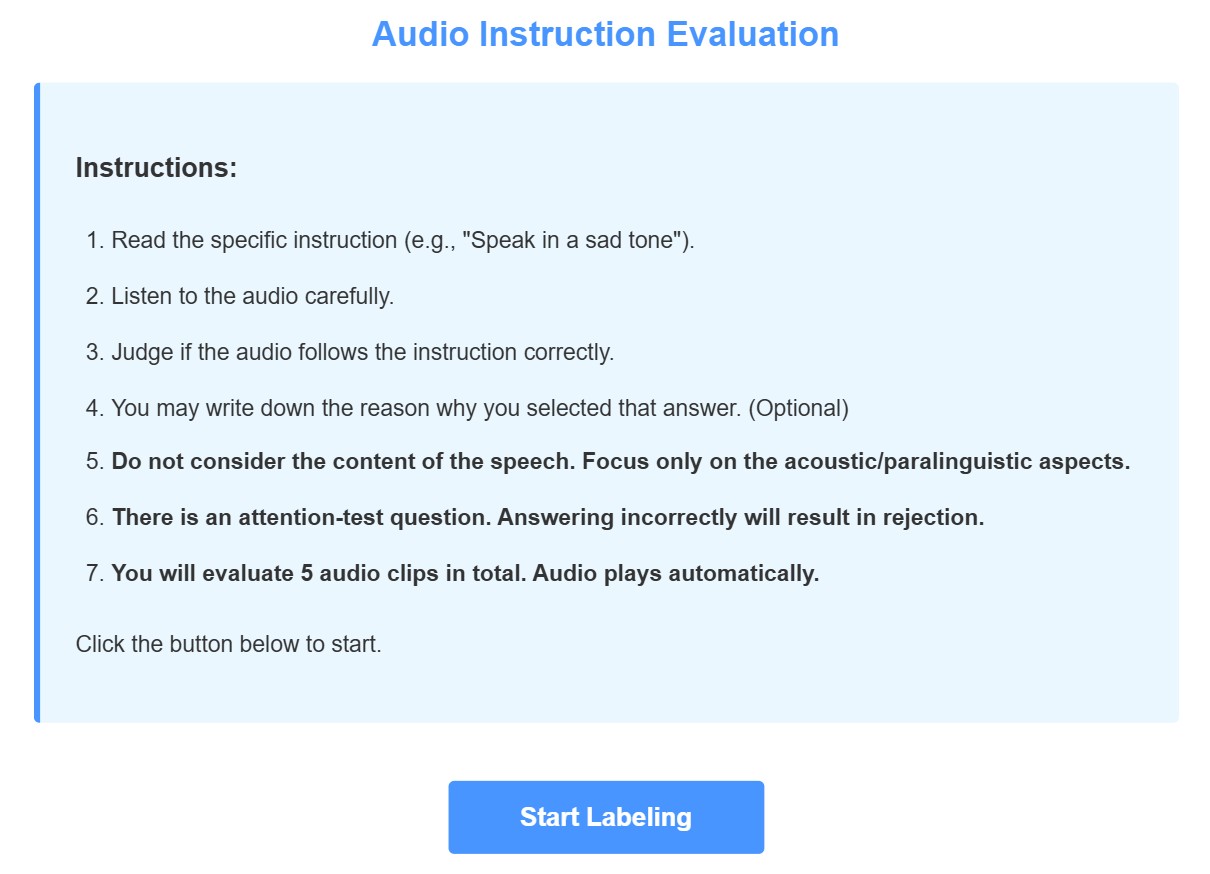}
    \caption{Instruction page}
    \label{fig:mturk_instruction_page}
    
    \vspace{1em}
    
    \includegraphics[width=0.85\textwidth]{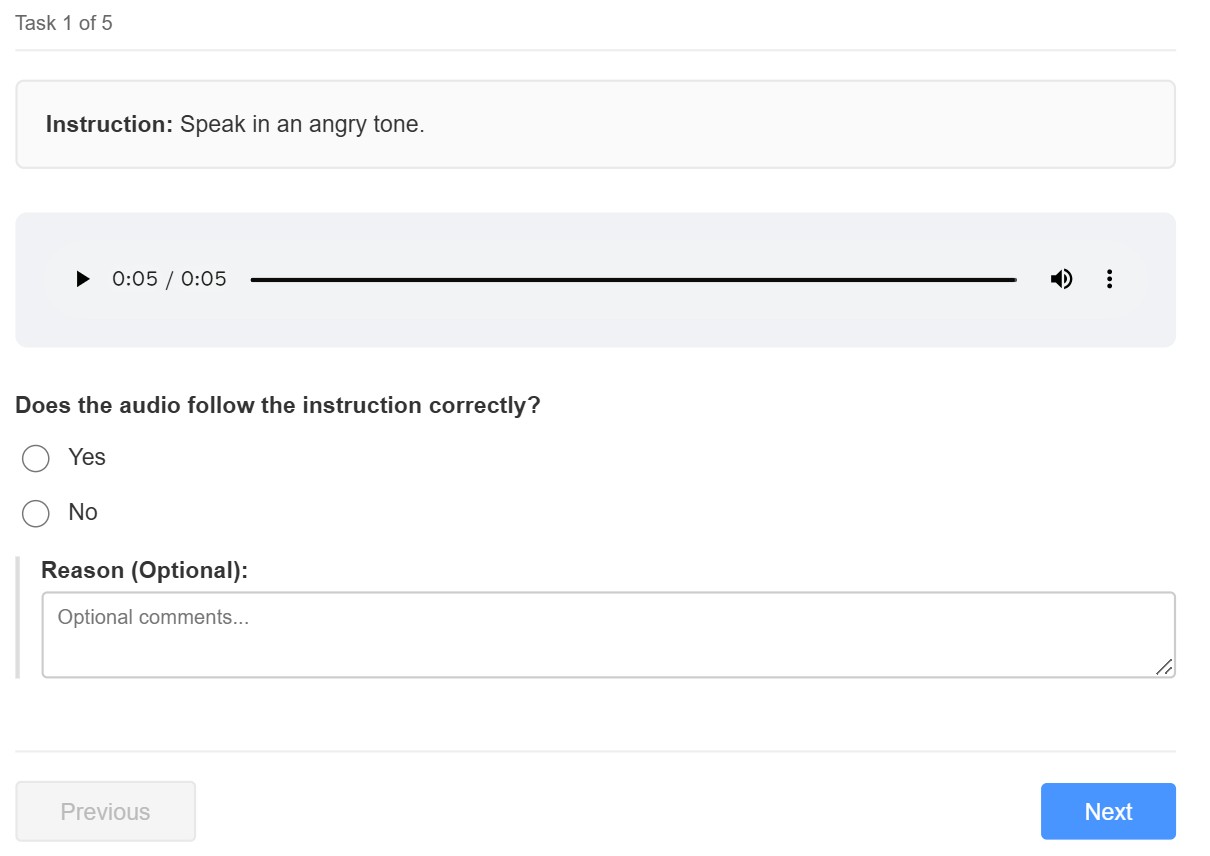}
    \caption{Annotation page}
    \label{fig:mturk_annotation_page}
\end{figure*}

In addition, we include an attention-check question to ensure that annotators actually listen to the audio before answering.
The attention-check is a short audio clip that clearly states the correct answer, so annotators must listen to the audio to respond correctly.
Each task contains four evaluation samples and one attention-check sample in random order.
If an annotator fails the attention check, we reject the submission and reassign the task to a new annotator.
The annotation interface is shown in Figure~\ref{fig:mturk_instruction_page} and Figure~\ref{fig:mturk_annotation_page}.

We pay each annotator \$0.15 per task to ensure that the payment meets minimum wage standards.

\subsection{Inter-Annotator Agreement of Speech Emotion Recognition Datasets}
\label{app:emotion_dataset_iia}
IEMOCAP~\cite{busso2008iemocap}, CREMA-D~\cite{cao2014crema}, and MELD~\cite{poria2019meld} are three widely used speech emotion recognition datasets.
These datasets provide large-scale speech emotion annotations through crowdsourcing, making significant contributions to the development of speech emotion recognition models.
Emotion in speech is inherently subjective and may be perceived differently by different annotators. 
As a result, inter-annotator agreement (IAA) is often imperfect.
This variability does not indicate annotation noise or low quality, but reflects the inherent subjectivity of emotion perception in speech.

Among them, IEMOCAP and MELD adopt Fleiss’ Kappa~\cite{fleiss1971measuring} as the IAA  indicator. 
Fleiss’ Kappa is designed to assess the reliability of agreement among a fixed number of annotators assigning categorical labels to a set of items, measuring the extent to which the observed agreement exceeds what would be expected by chance.
Each sample in the IEMOCAP dataset is annotated by three evaluators, yielding a Fleiss’ Kappa of 0.27, whereas the MELD dataset demonstrates a higher agreement of 0.43 among three annotators.

In comparison, the CREMA-D dataset uses  Krippendorff's alpha~\cite{krippendorff2018content} to report the agreement.
This metric is used to measure agreement across any number of observers and is particularly robust because it can handle incomplete data and various data scales while accounting for disagreements expected by chance.
The authors collect samples with an average of 9.8 annotators per clip and exhibit a Krippendorff's alpha of 0.42.
Furthermore, they report a self-consistency rate of approximately 70\%, which represents the level of agreement when the same annotator evaluates the same sample at different times.
In sum, these scores highlight the inherent subjectivity and complexity of emotional annotation in spoken dialogue.

\begin{figure*}[t]
    \begin{tcolorbox}[
      colback=gray!10,     
      colframe=gray!80,    
      title=Prompt for Dataset Construction, 
      fonttitle=\bfseries, 
      coltitle=white,      
      rounded corners,     
      width=\textwidth,    
      boxrule=0.5mm        
    ]
    \# Task \\
    You are given two pieces of information about a conversation:
    \\
    (1) A short narrative context that describes the social situation.
    \\
    (2) The original first utterance that started the dialogue.
    
    Rewrite the first utterance into a stronger, more natural opening line that better fits the narrative context.
    \\
    \\
    \# Guidelines\\
    - This sentence will be used as the initial input to start a conversation with an AI assistant. If the given conversation is not appropriate for interacting with an AI. For example, if it’s clearly directed toward a specific person, then respond only with ``no.''
    \\
    - The utterance should be open-ended, encouraging multi-turn, in-depth discussions rather than prompting a single, definitive response.
    \\
    - Keep the opener suitable for starting a conversation in this situation.
    \\
    - Preserve the core intent/topic of the original first utterance when appropriate, but improve clarity, grounding, and engagement.
    \\
    - You may slightly adjust the angle to better align with the narrative, but do NOT invent new facts beyond what the narrative implies.
    \\
    - Do not mention ``narrative'' or ``dialogue'' or that you are rewriting; just produce the line.
    \\
    - Output ONLY the rewritten opening line (no numbering, quotes, or extra text).
    \\
    \\
    \# Narrative: \\
    \{narrative in Soda\}
    \\
    \\
    \# Original first utterance: \\
    \{first utterance in Soda\}
    \end{tcolorbox}
    \caption{Prompt for dataset construction.}
    \label{fig:prompt_dataset}
\end{figure*}

\begin{figure*}[t]
    \begin{tcolorbox}[
      colback=gray!10,     
      colframe=gray!80,    
      title=Prompt for Dialogue Coherence Evaluation, 
      fonttitle=\bfseries, 
      coltitle=white,      
      rounded corners,     
      width=\textwidth,    
      boxrule=0.5mm        
    ]
    \# Task \\
    Your task is to evaluate the quality, coherence, and naturalness of a dialogue.
    The dialogue provided involves two participants: a ``Referee'' and a ``Participant''.
    \\ 
    \\ 
    Your job is to assess the **Participant's responses** to the ``Referee''. You must evaluate the naturalness, coherence, and overall reasonableness of the **Participant's replies only**. Do not score the Referee's sentence.
    \\
    Focus on whether the Participant's replies are logical, on-topic, and sound natural in the context of the conversation.
    \\ 
    \\
    \# Evaluation Steps \\
    1. ** Analyze the Dialogue Context ** 
    \\
    Read the entire dialogue history to understand the conversational flow. Identify the turns belonging to the 'Referee' and the 'Participant'. 
    \\
    2. ** Evaluate Participant's Responses **
    \\
    Review all responses made by the 'Participant'. Evaluate their overall quality based on the following criteria:
    \\
    - **Coherence:** 
    \\
    Are the replies logically connected to the Referee's statements? Do they make sense in context, or are they frequently off-topic? 
    \\
    - **Naturalness \& Reasonableness:** 
    \\
    Do the replies sound like a real person would say them? Is the content reasonable and appropriate? Do the responses show appropriate depth, or are they overly simplistic/robotic?
    \\
    3. ** Provide Analysis **
    \\
    Summarize your findings. Justify your final score by highlighting specific examples of good (coherent, natural) or poor (incoherent, unnatural) responses from the Participant.
    \\
    4. ** Report the Final Score **
    \\
    Conclude your evaluation with the following format: Final score: [[score]]. Replace score with an integer in \{\{score\_set\}\}. Keep the brackets as shown.
    \\
    \\
    \# Scoring Rubric
    \\
            - 1: **Completely Incoherent**: The Participant's replies are semantically unrelated to the Referee's statements. They are random, nonsensical, or completely off-topic.
            \\
            - 2: **Mostly Incoherent**: The Participant's replies are only vaguely related (e.g., catching a keyword but missing the point) or frequently introduce irrelevant topics, making the dialogue logically hard to follow.
            \\
            - 3: **Partially Coherent**: The Participant's replies are generally understandable and respond to the Referee, but contain clear logical leaps, topic drift, or semantic inconsistencies.
            \\
            - 4: **Mostly Coherent**: The Participant's replies are logical follow-ups and stay on-topic. The dialogue is semantically smooth, with only minor imprecision.
            \\
            - 5: **Highly Coherent**: The Participant's replies are semantically tightly-coupled to the Referee's statements, logically sound, and accurately advance the conversation, making it very fluent."
            \\
    \\
    
    \# Dialogue
    \\
    \{dialogue\}
    \end{tcolorbox}
    \caption{Prompt for dialogue coherence evaluation}
\label{fig:prompt_dialogue_coherence_evaluation}
\end{figure*}

\begin{figure*}[ht]
    \begin{tcolorbox}[
      colback=gray!10,     
      colframe=gray!80,    
      title=Prompt for Recall Evaluation, 
      fonttitle=\bfseries, 
      coltitle=white,      
      rounded corners,     
      width=\textwidth,    
      boxrule=0.5mm        
    ]
    \# User Instruction (Ground Truth):
    \\
    \{instruction\}
    \\
    \\
    \# Model Response: 
    \\
    \{response\}
    \\
    \\
    \# Question:
    \\
    Please evaluate the Model Response based on the User Instruction. Determine if the model correctly recalled the specific instruction given by the user.
    \\
    \\
    Select one of the following categories:
    \\
    (A) The response is not answering the Question, is unrelated, meaningless, or avoids the Question.
    \\
    (B) The response gives an instruction but different from the User Instruction (Ground Truth).
    \\
    (C) The response answers the question correctly but includes some meaningless sentences that are unrelated to the question.
    \\
    (D) The response answers and is completely correct regarding the User Instruction.
    \\
    \\
    Return only the single letter of the category (A, B, C, D).
    \end{tcolorbox}
    \caption{Prompt for recall evaluation}
    \label{fig:prompt_recall_evaluation}
\end{figure*}

\begin{table*}[t]
\centering

\begin{minipage}{0.48\textwidth}
\centering
\resizebox{\textwidth}{!}{
\begin{tabular}{llcccc}
\toprule
\multirow{2}{*}{\textbf{Style}} & \multirow{2}{*}{\textbf{Model}} & \multicolumn{4}{c}{\textbf{Assistant Turn}} \\ \cmidrule(lr){3-6}
 & & 1 & 2 & 3 & 4 \\ \midrule
\multirow{6}{*}{\textbf{Anger}} & Cascaded Baseline & 17.0 & 14.0 & 17.0 & 13.0 \\
\cmidrule{2-6}
 & Gemini Live & 24.0 & 12.0 & 10.0 & 8.0 \\
 & GPT-4o & 30.5 & 24.2 & 17.9 & 8.4 \\
 & GPT-4o mini & 39.0 & 6.0 & 6.0 & 1.0 \\
 & Step-Audio 2 mini & 1.0 & 0.0 & 0.0 & 0.0 \\
 & Qwen2.5-Omni & 5.0 & 1.0 & 0.0 & 3.0 \\ \midrule
\multirow{6}{*}{\textbf{Happiness}} & Cascaded Baseline & 85.0 & 79.0 & 87.0 & 82.0 \\
\cmidrule{2-6}
 & Gemini Live & 91.0 & 89.0 & 89.0 & 92.0 \\
 & GPT-4o & 82.1 & 92.6 & 87.4 & 85.3 \\
 & GPT-4o mini & 89.0 & 90.0 & 87.0 & 84.0 \\
 & Step-Audio 2 mini & 100.0 & 99.0 & 98.0 & 97.0 \\
 & Qwen2.5-Omni & 71.0 & 87.0 & 83.0 & 77.0 \\ \midrule
\multirow{6}{*}{\textbf{Neutral}} & Cascaded Baseline & 73.0 & 79.0 & 75.0 & 74.0 \\
\cmidrule{2-6}
 & Gemini Live & 64.0 & 69.0 & 75.0 & 69.0 \\
 & GPT-4o & 58.8 & 62.9 & 59.8 & 56.7 \\
 & GPT-4o mini & 45.0 & 35.0 & 40.0 & 27.0 \\
 & Step-Audio 2 mini & 7.0 & 3.0 & 3.0 & 5.0 \\
 & Qwen2.5-Omni & 41.0 & 20.0 & 17.0 & 18.0 \\ \midrule
\multirow{6}{*}{\textbf{Sadness}} & Cascaded Baseline & 62.0 & 58.0 & 60.0 & 64.0 \\
\cmidrule{2-6}
 & Gemini Live & 72.0 & 54.0 & 47.0 & 51.0 \\
 & GPT-4o & 78.0 & 65.0 & 44.0 & 45.0 \\
 & GPT-4o mini & 85.0 & 32.0 & 18.0 & 9.0 \\
 & Step-Audio 2 mini & 17.0 & 4.0 & 4.0 & 1.0 \\
 & Qwen2.5-Omni & 17.0 & 4.0 & 4.0 & 0.0 \\ \bottomrule
\end{tabular}}
\end{minipage}%
\hfill
\begin{minipage}{0.48\textwidth}
\centering
\resizebox{\textwidth}{!}{
\begin{tabular}{llcccc}
\toprule
\multirow{2}{*}{\textbf{Style}} & \multirow{2}{*}{\textbf{Model}} & \multicolumn{4}{c}{\textbf{Assistant Turn}} \\ \cmidrule(lr){3-6}
 & & 1 & 2 & 3 & 4 \\ \midrule
\multirow{6}{*}{\textbf{North American}}
 & Cascaded Baseline & 100.0 & 99.0 & 100.0 & 100.0 \\
\cmidrule{2-6}
 & Gemini Live & 100.0 & 100.0 & 100.0 & 100.0 \\
 & GPT-4o & 100.0 & 100.0 & 100.0 & 100.0 \\
 & GPT-4o mini & 100.0 & 100.0 & 100.0 & 100.0 \\
 & Step-Audio 2 mini & 66.0 & 78.0 & 72.0 & 71.0 \\
 & Qwen2.5-Omni & 100.0 & 99.0 & 100.0 & 100.0 \\ \midrule
\multirow{6}{*}{\textbf{Indian}}
 & Cascaded Baseline & 100.0 & 100.0 & 99.0 & 100.0 \\
\cmidrule{2-6}
 & Gemini Live & 100.0 & 100.0 & 100.0 & 100.0 \\
 & GPT-4o & 100.0 & 100.0 & 98.0 & 97.0 \\
 & GPT-4o mini & 89.0 & 57.0 & 35.0 & 26.0 \\
 & Step-Audio 2 mini & 33.0 & 21.0 & 36.0 & 30.0 \\
 & Qwen2.5-Omni & 0.0 & 0.0 & 0.0 & 0.0 \\ \bottomrule
\end{tabular}}
\end{minipage}

\vspace{1em}


\begin{minipage}{0.48\textwidth}
\centering
\resizebox{\textwidth}{!}{
\begin{tabular}{llcccc}
\toprule
\multirow{2}{*}{\textbf{Style}} & \multirow{2}{*}{\textbf{Model}} & \multicolumn{4}{c}{\textbf{Assistant Turn}} \\ \cmidrule(lr){3-6}
 & & 1 & 2 & 3 & 4 \\ \midrule
\multirow{6}{*}{\textbf{Loud}} 
 & Cascaded Baseline & 96.0 & 97.0 & 98.0 & 99.0 \\
\cmidrule{2-6}
 & Gemini Live & 57.0 & 55.0 & 60.0 & 73.0 \\
 & GPT-4o & 67.0 & 68.0 & 59.0 & 56.0 \\
 & GPT-4o mini & 77.0 & 74.0 & 68.0 & 61.0 \\
 & Step-Audio 2 mini & 46.0 & 50.0 & 44.0 & 49.0 \\
 & Qwen2.5-Omni & 38.0 & 36.0 & 36.0 & 41.0 \\ \midrule
\multirow{6}{*}{\textbf{Quiet}}
 & Cascaded Baseline & 99.0 & 99.0 & 97.0 & 97.0 \\
\cmidrule{2-6}
 & Gemini Live & 95.0 & 93.0 & 88.0 & 82.0 \\
 & GPT-4o & 92.7 & 94.8 & 94.8 & 95.8 \\
 & GPT-4o mini & 100.0 & 99.0 & 99.0 & 99.0 \\
 & Step-Audio 2 mini & 56.0 & 66.0 & 52.0 & 51.0 \\
 & Qwen2.5-Omni & 69.0 & 63.0 & 64.0 & 59.0 \\ \bottomrule
\end{tabular}}
\end{minipage}%
\hfill
\begin{minipage}{0.48\textwidth}
\centering
\resizebox{\textwidth}{!}{
\begin{tabular}{llcccc}
\toprule
\multirow{2}{*}{\textbf{Style}} & \multirow{2}{*}{\textbf{Model}} & \multicolumn{4}{c}{\textbf{Assistant Turn}} \\ \cmidrule(lr){3-6}
 & & 1 & 2 & 3 & 4 \\ \midrule
\multirow{6}{*}{\textbf{Fast}}
 & Cascaded Baseline & 100.0 & 100.0 & 99.0 & 99.0 \\
\cmidrule{2-6}
 & Gemini Live & 99.0 & 97.0 & 86.0 & 85.0 \\
 & GPT-4o & 89.0 & 87.0 & 90.0 & 81.0 \\
 & GPT-4o mini & 87.0 & 86.0 & 83.0 & 80.0 \\
 & Step-Audio 2 mini & 89.0 & 64.0 & 60.0 & 62.0 \\
 & Qwen2.5-Omni & 47.0 & 49.0 & 24.0 & 27.0 \\ \midrule
\multirow{6}{*}{\textbf{Slow}}
 & Cascaded Baseline & 100.0 & 100.0 & 98.0 & 100.0 \\
\cmidrule{2-6}
 & Gemini Live & 99.0 & 99.0 & 98.0 & 98.0 \\
 & GPT-4o & 100.0 & 96.0 & 92.9 & 86.9 \\
 & GPT-4o mini & 99.0 & 83.8 & 80.8 & 69.7 \\
 & Step-Audio 2 mini & 81.0 & 67.0 & 53.0 & 42.0 \\
 & Qwen2.5-Omni & 40.0 & 66.0 & 71.0 & 76.0 \\ \bottomrule
\end{tabular}}
\end{minipage}

\caption{IF rate across emotion, accent, volume, and speed styles.}
\label{tab:full_style_degradation}
\end{table*}

\begin{table*}
    \centering 
    \begin{tabular}{l | ccccc} 
        \toprule
        \textbf{Style} & \textbf{Gemini~Live} & \textbf{GPT-4o} & \textbf{GPT-4o~mini} & \textbf{Qwen2.5-Omni} & \textbf{Step-Audio 2 mini} \\
        \midrule
        Anger & 3.80 & 4.19 & 4.13 & 4.09 & 3.53 \\
        Happiness & 4.18 & 4.20 & 4.22 & 4.23 & 3.60 \\
        Sadness & 4.31 & 4.28 & 4.20 & 4.17 & 3.49 \\
        Neutral & 4.28 & 4.21 & 4.35 & 4.18 & 3.59 \\
        Fast & 4.27 & 4.26 & 4.18 & 4.09 & 3.51 \\
        Slow & 4.09 & 4.22 & 4.04 & 4.12 & 3.50 \\
        Loud & 4.00 & 4.19 & 4.17 & 4.29 & 3.49 \\
        Quiet & 4.18 & 4.23 & 4.19 & 4.05 & 3.59 \\
        Indian & 4.17 & 4.22 & 4.25 & 4.36 & 3.54 \\
        North American & 4.20 & 4.18 & 4.16 & 4.17 & 3.61 \\
        \bottomrule
    \end{tabular}
    
    \caption{Dialogue coherence evaluation results.} 
    \label{tab:semantic_evaluation}
\end{table*}

\begin{table*}
    \centering
    \begin{tabular}{l|cccc}
        \toprule
        \textbf{Style} & \textbf{GPT-4o} & \textbf{GPT-4o mini} & \textbf{GPT-4o + recall} & \textbf{GPT-4o mini + recall} \\
        \midrule
        Anger & 95 & 100 & - & - \\
        Happiness & 95 & 100 & - & - \\
        Sadness & 100 & 100 & 85 & 100 \\
        Neutral  & 97 & 100 & - & - \\
        Fast  & 100 & 99 & 99 & 99 \\
        Slow  & 99 & 89 & 99 & 89 \\
        Loud  & 100 & 100 & - & - \\
        Quiet  & 96 & 100 & - & - \\
        Indian  & 100 & 100 & 86 & 96 \\
        North American  & 100 & 100 & - & - \\
        \bottomrule
    \end{tabular}
    \caption{The number of successfully generated samples after up to three retries}
    \label{tab:gpt_successful_number}
\end{table*}

\begin{table*}[ht]
    \centering
    \label{tab:prompt_position_ifrate_one_decimal}
    \resizebox{0.8\textwidth}{!}{
    \begin{tabular}{llcrrrrr}
    \toprule
    \multirow{2}{*}{\textbf{Model}} & \multirow{2}{*}{\textbf{Style}} & \multirow{2}{*}{\textbf{Position}} & \multicolumn{4}{c}{\textbf{Turn}} \\
    \cmidrule(lr){4-7}
    & & & \multicolumn{1}{c}{1} & \multicolumn{1}{c}{2} & \multicolumn{1}{c}{3} & \multicolumn{1}{c}{4} \\
    \midrule
    \multirow{10}{*}{GPT-4o} & \multirow{2}{*}{Anger} & System & 5.0 & 4.0 & 1.0 & 0.0 \\
    & & User & \textbf{30.5} & \textbf{24.2} & \textbf{17.9} & \textbf{8.4} \\
    \cmidrule{2-7}
    & \multirow{2}{*}{Sadness} & System & 39.0 & 18.2 & 11.1 & 12.2 \\
    & & User & \textbf{78.0} & \textbf{65.0} & \textbf{44.0} & \textbf{45.0} \\
    \cmidrule{2-7}
    & \multirow{2}{*}{Indian} & System & 50.0 & 24.0 & 11.8 & 5.6 \\
    & & User & \textbf{100.0} & \textbf{100.0} & \textbf{98.0} & \textbf{97.0} \\
    \cmidrule{2-7}
    & \multirow{2}{*}{Fast} & System & 62.0 & 61.0 & 61.6 & 62.1 \\
    & & User & \textbf{89.0} & \textbf{87.0} & \textbf{90.0} & \textbf{81.0} \\
    \cmidrule{2-7}
    & \multirow{2}{*}{Slow} & System & 56.0 & 45.5 & 38.8 & 39.5 \\
    & & User & \textbf{100.0} & \textbf{96.0} & \textbf{92.9} & \textbf{86.9} \\

    \midrule
    \multirow{10}{*}{GPT-4o mini} & \multirow{2}{*}{Anger} & System & 3.0 & 0.0 & 0.0 & 0.0 \\
    & & User & \textbf{39.0} & \textbf{6.0} & \textbf{6.0} & \textbf{1.0} \\
    \cmidrule{2-7}
    & \multirow{2}{*}{Sadness} & System & 33.0 & 8.0 & 4.0 & 2.0 \\
    & & User & \textbf{85.0} & \textbf{32.0} & \textbf{18.0} & \textbf{9.0} \\
    \cmidrule{2-7}
    & \multirow{2}{*}{Indian} & System & 0.0 & 1.0 & 0.0 & 2.0 \\
    & & User & \textbf{89.0} & \textbf{57.0} & \textbf{35.0} & \textbf{26.0} \\
    \cmidrule{2-7}
    & \multirow{2}{*}{Fast} & System & 54.0 & 53.0 & 63.0 & 62.0 \\
    & & User & \textbf{87.0} & \textbf{86.0} & \textbf{83.0} & \textbf{80.0} \\
    \cmidrule{2-7}
    & \multirow{2}{*}{Slow} & System & 62.0 & 55.0 & 44.0 & 47.0 \\
    & & User & \textbf{99.0} & \textbf{83.8} & \textbf{80.8} & \textbf{69.7} \\

    \midrule
    \multirow{10}{*}{Step-Audio 2 mini} & \multirow{2}{*}{Anger} & System & 0.0 & 0.0 & 0.0 & 0.0 \\
    & & User & \textbf{1.0} & 0.0 & 0.0 & 0.0 \\
    \cmidrule{2-7}
    & \multirow{2}{*}{Sadness} & System & 4.0 & 3.0 & 0.0 & \textbf{1.0} \\
    & & User & \textbf{17.0} & \textbf{4.0} & \textbf{4.0} & \textbf{1.0} \\
    \cmidrule{2-7}
    & \multirow{2}{*}{Indian} & System & \textbf{32.0} & \textbf{25.0} & 28.0 & \textbf{30.0} \\
    & & User & \textbf{32.0} & 21.0 & \textbf{33.0} & 28.0 \\
    \cmidrule{2-7}
    & \multirow{2}{*}{Fast} & System & 79.0 & \textbf{69.0} & \textbf{65.0} & \textbf{73.0} \\
    & & User & \textbf{89.0} & 64.0 & 60.0 & 62.0 \\
    \cmidrule{2-7}
    & \multirow{2}{*}{Slow} & System & 50.0 & 55.0 & 51.0 & \textbf{50.0} \\
    & & User & \textbf{81.0} & \textbf{67.0} & \textbf{53.0} & 42.0 \\
    \bottomrule
    \end{tabular}%
    }
    \caption{IF rate comparison by prompt position. \textbf{Bold} indicates better performance between user and system messages.}
    \label{tab:full_prompt_position}
\end{table*}

\end{document}